\documentclass[10pt,twocolumn]{article}

\usepackage[pagenumbers]{cvpr} 

\usepackage[dvipsnames]{xcolor}
\definecolor{isabelline}{rgb}{0.96, 0.94, 0.93}
\definecolor{LightYellow}{rgb}{1.0, 1.0, 0.88}
\definecolor{magicmint}{rgb}{0.67, 0.94, 0.82}
\definecolor{lightmauve}{rgb}{0.86, 0.82, 1.0}
\definecolor{grannysmithapple}{rgb}{0.66, 0.89, 0.63}
\definecolor{coralpink}{rgb}{0.97, 0.51, 0.47}
\definecolor{cottoncandy}{rgb}{1.0, 0.74, 0.85}

\usepackage[accsupp]{axessibility} 

\usepackage{times}
\usepackage{epsfig}
\usepackage{graphicx}
\usepackage{amsmath}
\usepackage{amssymb}
\usepackage{algorithm}
\usepackage{algorithmic}
\usepackage{multirow}
\usepackage{array}
\usepackage{booktabs}
\usepackage{makecell}
\usepackage{float}
\usepackage{diagbox}
\usepackage{changepage}
\usepackage{cuted} 
\usepackage{xcolor}
\usepackage{caption}
\usepackage{subcaption}
\usepackage{dsfont}
\usepackage{chngcntr}
\usepackage{cuted}
\usepackage{capt-of}
\usepackage{color, colortbl}
\definecolor{LightCyan}{rgb}{0.88,1,1}

\definecolor{cvprblue}{rgb}{0.21,0.49,0.74}
\usepackage[pagebackref,breaklinks,colorlinks,citecolor=cvprblue]{hyperref}

\def\paperID{2319} 
\def\confName{CVPR}
\def\confYear{2024}

\usepackage[capitalize]{cleveref}
\crefname{section}{Sec.}{Secs.}
\Crefname{section}{Section}{Sections}
\Crefname{table}{Table}{Tables}
\crefname{table}{Table}{Tables}

\usepackage{tikz}
\newcommand\crossmark[1][]{%
  \tikz[scale=0.25,#1]{
    \fill(0,0)--(0.1,0) .. controls (0.5,0.4) .. (1,0.7)--(0.9,0.7) ..  controls (0.5,0.5) ..(0,0.1) --cycle;
    \fill(1,0.1)--(0.9,0.1) .. controls (0.5,0.3) .. (0,0.7)--(0.1,0.7) .. controls (0.5,0.4) ..(1,0.2) --cycle;
  }%
}

\begin{document}

\title{UniVS: Unified and Universal Video Segmentation with Prompts as Queries}

\author{Minghan Li$^{1,2}$\thanks{Equal contribution, $\dagger$ Corresponding author.}, Shuai Li$^{1,2*}$, Xindong Zhang$^2$ and Lei Zhang$^{1,2\dagger}$  \\
$^1$The Hong Kong Polytechnic University \quad $^2$OPPO Research Institute \\
{\tt\small liminghan0330@gmail.com}, 
{\tt\small xindongzhang@foxmail.com},
{\tt\small \{csshuaili, cslzhang\}@comp.polyu.edu.hk}
}

\maketitle

\begin{strip}
\centering
\includegraphics[width=0.98\linewidth]{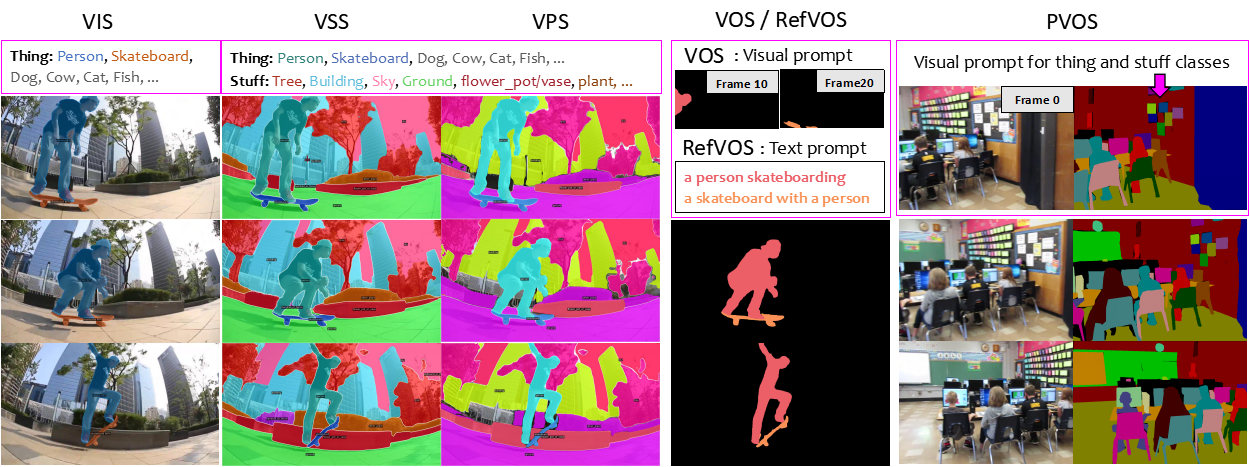}
\vspace{-1mm}
\captionof{figure}{Illustration of different video segmentation (VS) tasks.  Category-specified VS includes VIS, VSS and VPS tasks, while prompt-specified VS consists of VOS, RefVOS and PVOS tasks. Please find more video demos on our project page \url{https://sites.google.com/view/unified-video-seg-univs}. }\label{fig:vs_tasks}
\vspace{+2mm}
\end{strip}

\begin{abstract}
Despite the recent advances in unified image segmentation (IS), developing a unified video segmentation (VS) model remains a challenge. This is mainly because generic category-specified VS tasks need to detect all objects and track them across consecutive frames, while prompt-guided VS tasks require re-identifying the target with visual/text prompts throughout the entire video, making it hard to handle the different tasks with the same architecture. We make an attempt to address these issues and present a novel unified VS architecture, namely UniVS, by using prompts as queries. UniVS averages the prompt features of the target from previous frames as its initial query to explicitly decode masks, and introduces a target-wise prompt cross-attention layer in the mask decoder to integrate prompt features in the memory pool. By taking the predicted masks of entities from previous frames as their visual prompts, UniVS converts different VS tasks into prompt-guided target segmentation, eliminating the heuristic inter-frame matching process. Our framework not only unifies the different VS tasks but also naturally achieves universal training and testing, ensuring robust performance across different scenarios. UniVS shows a commendable balance between performance and universality on 10 challenging VS benchmarks, covering video instance, semantic, panoptic, object, and referring segmentation tasks. Code can be found at \url{https://github.com/MinghanLi/UniVS}.

\end{abstract} 
\section{Introduction}
Video segmentation (VS) partitions a video sequence into different regions or segments, facilitating many applications such as semantic-guided video restoration \cite{li2018video,wu2024seesr,sun2023improving}, video generation \cite{Peebles2022DiT,rombach2021highresolution}, video editing and augmented reality \cite{wei2023elite,zhang2023controlvideo,guo2023animatediff}, \etc.
VS tasks can be divided into two groups: category-specified VS and prompt-specified VS. The former focuses on segmenting and tracking entities from a predefined set of categories. Typical tasks along this line include video instance \cite{yang2019video,qi2021occluded}, semantic \cite{vspw} and panoptic segmentation \cite{vps,vipseg} (VIS/VSS/VPS), where the object category information needs to be specified. Another group focuses on identifying and segmenting specific targets throughout the video, where visual prompts or textual descriptions of the targets need to be provided. 
Prompt-specified VS tasks include video object segmentation (VOS) \cite{Perazzi2016vos,youtubevos,mose,burst}, Panoptic VOS (PVOS) \cite{xu2023viposeg} and referring VOS (RefVOS) \cite{refvos}. Each of above VS tasks has established its own protocol for dataset annotation, as well as model evaluation. 

VS has a close relationship with image segmentation (IS) \cite{everingham2010pascalvoc,lin2014microsoft,ade20k,panoptic_seg,cordts2016cityscapes,xu2016deep_interactive_seg,extreme_interactive_seg,refcoco,zhou2023interseg-GP,peng2022semantic}, which has similar types of segmentation tasks to VS. 
In the past decades, the model performance on each individual IS/VS task has been significantly improved, and many well-known network architectures have been developed  \cite{he2017mask,bolya2019yolact,chen2017deeplab,fcn_ss,2015Unet,deeperlab_ps,panoptic_deeplab,extreme_interactive_seg,refer,grounding_refer,Li_2021_CVPR,wang2020vistr,xmem,IDOL,huang2022minvis,openvis}. Some architectures \cite{k-net,cheng2021mask2former, li2022maskdino,zou2022xdecoder, videoknet, videokmax, tubelink, dvis} have also been proposed to deal with multiple segmentation tasks; however, these architectures require separate training and inference for different tasks because of the different annotations and evaluation protocols. 
Fortunately, the recent advancements in vision-language models \cite{li2022blip,clip,12-in-1,li2021glip,regionclip,jin2023context,zhang2024dual,liu2023llava,lin2023videollava,liu2023LargeWorldModel,yu2023magvit} align and harmonize the multimodal feature representations, which bridge the labeling gaps across various IS/VS tasks. As a result, some unified segmentation models \cite{freeseg, seem, semanticsam} have emerged to process multiple segmentation tasks simultaneously, which can be jointly trained on different datasets and tasks. These methods generally convert prompt-guided segmentation into the category-specified segmentation problem \cite{he2017mask,zhu2020deformable_detr,cheng2021mask2former}, which first predicts masks for all potential entities per frame and then utilizes a post-processing matching to find the target. For example, SEEM \cite{seem} captures prompt information of the target via concatenating learnable queries and prompt features as keys and values in the self-attention layer of the mask decoder, showing good versatility in various IS tasks. 

\begin{figure}[t]
     \includegraphics[width=1.0\linewidth]{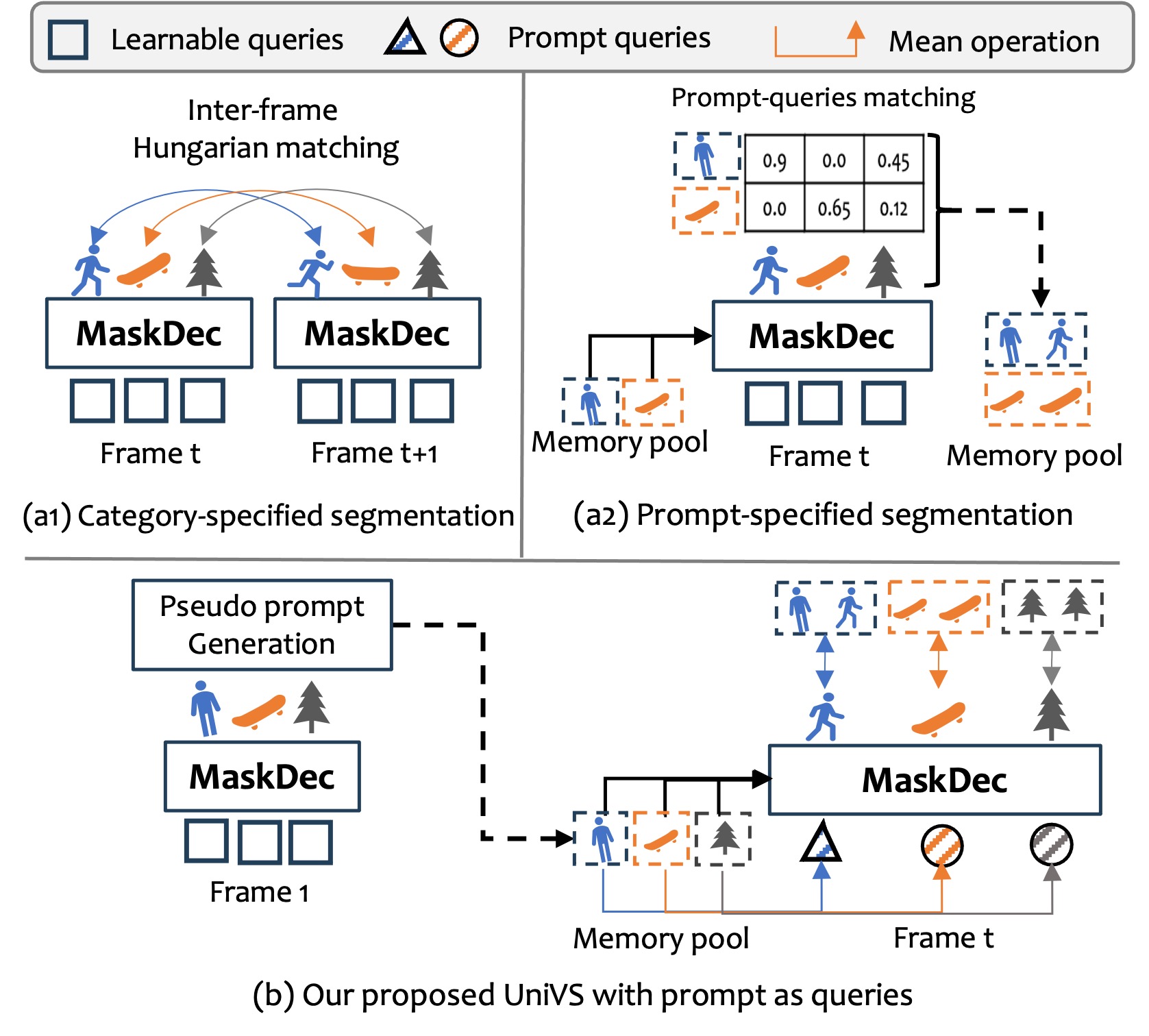}
     \vspace{-6mm}
     \caption{
     Comparison between the existing unified segmentation methods and ours. In existing methods for category-specified segmentation tasks (see (a1)), entities need to be first detected per frame and then matched across frames, while in methods for prompt-specified segmentation tasks (see (a2)), targets need to be identified from the predicted masks. In contrast, our proposed UniVS (see (b)) uses predicted masks as pseudo visual prompts and averages prompt features to decode masks across videos, avoiding the heuristic post-processing process.
     }\label{fig:prompt_as_queries}
     \vspace{-3mm}
\end{figure}

Compared to IS, the additional challenge of VS lies in the temporal consistency \cite{wang2021unitracker} of segmentation across frames in a video sequence. Existing unified models \cite{uninext,tarvis} for VS tasks are mostly inspired by the unified IS models. They segment the video sequence frame by frame, and then use a similarity matching step to associate common objects or find the targets for category-specified and prompt-specified VS tasks, respectively. For example, UNINEXT \cite{uninext} is specially designed for object-centric IS and VS tasks. Though UNINEXT performs very well in several aspects, it becomes ineffective when segmenting `stuff’ entities, such as `sky’. To adapt to different VS tasks, TarVIS \cite{tarvis} subdivides the learnable queries within the mask decoder into four groups: semantic, class-agnostic instance, object and background queries. However, TarVIS falls short in encoding linguistic information, hindering it from resolving language-guided VS tasks.

From the above discussions, we can see that it remains a challenge to develop a unified VS framework to effectively handle all VS tasks. This is mainly because category-specified and prompt-specified VS tasks attribute to different focuses. As shown in Figs. \ref{fig:prompt_as_queries}(a1) and \ref{fig:prompt_as_queries}(a2), category-specified segmentation prioritizes the precise detection per frame and the inter-frame association for common objects, while prompt-specified segmentation concentrates on accurately tracking the target with text/visual prompts in video sequences, where the target can be an uncommon object or a part of an object. The different focuses of the two types of VS tasks make it challenging to integrate them within a single framework while achieving satisfactory results.

To alleviate the above issues, we propose a novel unified VS architecture, namely \textbf{UniVS}, by using prompts as queries. For each target of interest, UniVS averages the prompt features from previous frames as its initial query. A target-wise prompt cross-attention (ProCA) layer is introduced in the mask decoder to integrate comprehensive prompt features stored in the memory pool. The initial query and the ProCA layer play a crucial role in the explicit and accurate decoding of masks.
On the other hand, by taking the predicted masks of entities from previous frames as their visual prompts, UniVS can convert different VS tasks into the task of prompt-guided target segmentation task, eliminating the heuristic inter-frame matching. The overall process of UniVS is depicted in Fig. \ref{fig:prompt_as_queries}(b). UniVS not only unifies the different VS tasks (see Fig. \ref{fig:vs_tasks}) but also naturally achieves universal training and testing, resulting in robust performance across different scenarios. It shows a commendable balance between performance and universality on 10 challenging VS benchmarks, covering VIS, VSS, VPS, VOS, RefVOS and PVOS tasks. To the best of our knowledge, \textit{\textbf{UniVS is the first work which can unify all the existing VS tasks successfully in a single model}}. 

\begin{figure*}[t]
     \vspace{-1mm}
     \centering
     \includegraphics[width=0.95\linewidth]{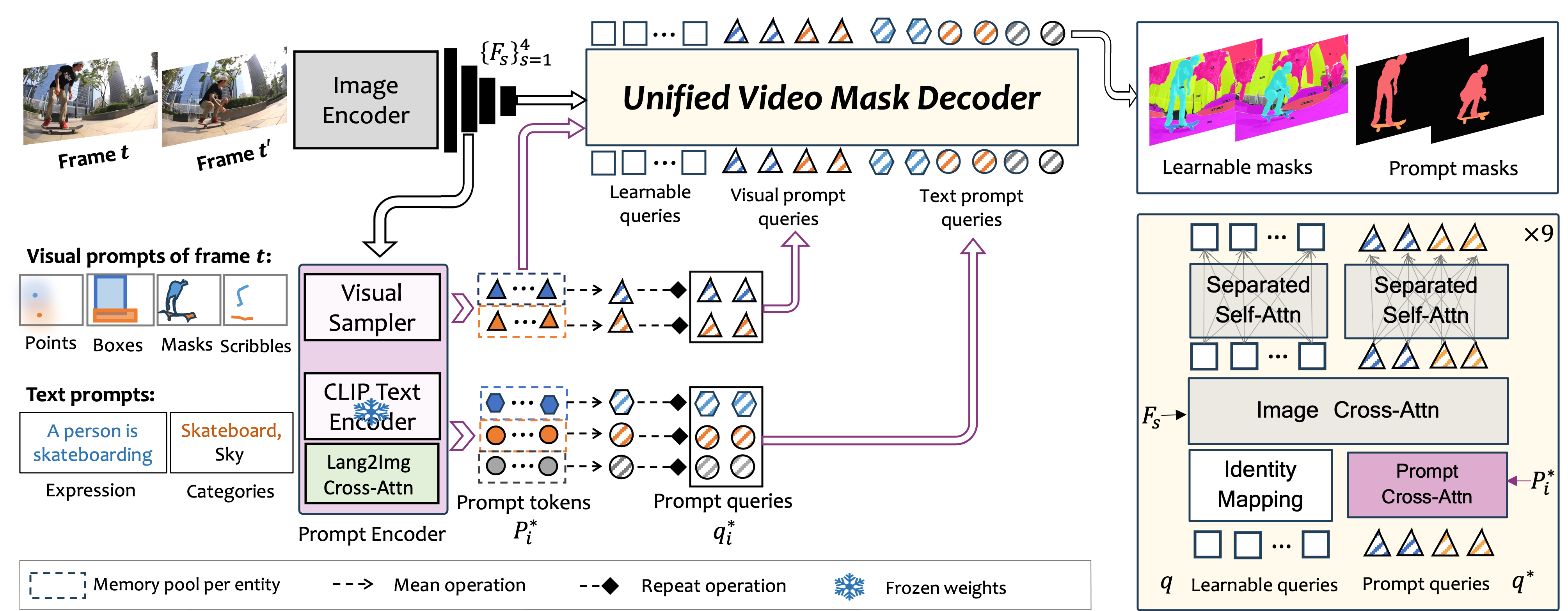}
     \vspace{-1mm}
     \caption{Training process of our unified video segmentation (UniVS) framework. UniVS contains three main modules: the Image Encoder (grey rectangle), the Prompt Encoder (purple rectangle) and the Unified Video Mask Decoder (yellow rectangle). The Image Encoder transforms the input RGB images to the feature space and outputs image embeddings. Meanwhile, the Prompt Encoder translates the raw visual/text prompts into prompt embeddings. The Unified Video Mask Decoder explicitly decodes the masks for any entity or prompt-guided target in the input video by using prompts as queries (striped triangles, hexagons and circles). }\label{fig:overview}
     \vspace{-2mm}
\end{figure*}
\section{Related Work}
We first briefly introduces the representative models designed for category-specified and prompt-specified VS tasks, and then introduce the existing unified and universal models for processing multiple VS tasks simultaneously.

\textbf{Category-specified VS} tasks includes video instance \cite{yang2019video}, semantic \cite{vspw} and panoptic segmentation \cite{vipseg} tasks, which aim to correctly partition the regions and label them with specific categories or open-world vocabulary\cite{openseed,li2023opensd,openvis}. Building on representative query-based image detection and segmentation methods \cite{carion2020detr,zhu2020deformable_detr,lishuai2023one,cheng2021mask2former}, recent approaches \cite{Li_2021_CVPR,wang2020vistr,seqformer,IDOL,heo2022vita,huang2022minvis,li2023mdqe,GenVIS,tubelink,tubeformer,videoknet} focus on efficiently encoding both short-term and medium-term temporal information. Some methods \cite{seqformer, IDOL,huang2022minvis,li2023mdqe,panoptic_fcn} utilize the consistency of the relative relationship between objects in a short period of time to associate entities across frames, while the latest state-of-the-arts \cite{heo2022vita,GenVIS,videoknet,tubelink,dvis} design learnable trackers between multiple video clips to better learn entity motion information over short to medium time periods, thus maintaining more temporally consistent segmentation results. This continual learning strategy also bridges the gap between training and inference.

\textbf{Prompt-specified VS} tasks consist of video object segmentation (VOS) \cite{youtubevos,davis,mose,burst}, panoptic VOS \cite{xu2023viposeg} and referring VOS \cite{refvos}, which aim to re-identify and segment the target objects with visual/text prompts in a video.
Recent offline models \cite{xmem,yang2022deaot,referformer,grounding_refer,li2022r2vos,Miao_2023_ICCV_sgmg,cheng2023cutie} focus more on designing long-term information propagation modules to transfer previous image features and corresponding masks to the target frame to predict masks. This helps more precisely identify and track the movement trajectory of objects throughout the entire video sequence (more than 200 frames), but also makes the network architecture heavy \cite{yang2022deaot,Wu_2023_ICCV_uniref}.
Due to the slow speed of offline models, online models \cite{wu2023onlinerefer,jointformer_vos,Wu_2023_ICCV_uniref} come back into focus, aiming to keep the right balance between speed and performance.

\textbf{Unified VS} models \cite{cheng2021mask2former-video,videoknet,videokmax,tubelink,dvis,uninext,tarvis,cheng2023DEVA_tracking,Wu_2023_ICCV_uniref} have emerged recently.
TubeLink \cite{tubelink} and DVIS \cite{dvis} employ a single framework to accomplish different category-specified VS tasks, but they still need to be trained separately for each task and cannot be used for prompt-specified VS tasks.
UniRef \cite{Wu_2023_ICCV_uniref} uses a model to unify prompt-specified VS tasks, but it cannot handle category-specified VS tasks. To handle more types of VS tasks, TarVIS \cite{tarvis} decouples learnable queries of mask decoder into four groups. But, it fails to solve RefVOS due to the lack of language encoding capability. UNINEXT \cite{uninext} is a unified object-centric segmentation model on images and videos, demonstrating universal applicability in several aspects. Unfortunately, it cannot handle entities with `stuff' categories.

By far, there still lacks a unified model that can accommodate all VS tasks simultaneously. This is mainly because different VS tasks have different focuses, which can be contradictory. To address this difficulty, we use prompts as queries to unify different tasks into a unified framework.

\section{Methodology}
UniVS contains of three main modules: the Image Encode, the Prompt Encoder and the Unified Video Mask Decoder, as depicted in Fig. \ref{fig:overview}. The Image Encoder transforms RGB images to the feature tokens, while the Prompt Encoder translates raw visual/text prompts into the prompt embeddings. The Unified Video Mask Decoder explicitly decodes masks for any entity or prompt-guided target in the video.

\subsection{Image and Prompt Encoders}

{\bf Image encoder} contains a backbone and a pixel decoder \cite{cheng2021mask2former}. The backbone maps the RGB image $X \in R^{3\times H\times W}$ into multi-scale features, and the pixel decoder further fuses features across scales to enhance the representation: 
$\{F_s\}_{s=1}^4 = \text{ImEnc}(X),$
where $F_s\!\in\! R^{C\times H_s \times W_s}$ is the $s$-th scale image embeddings, $H_s, W_s$ are its height and width and $C$ is the number of channels. The resolutions of multi-scale feature maps are 1/32, 1/16, 1/8 and 1/4 of that of the input image, respectively.

{\bf Prompt encoder} converts the input visual/text prompts into the prompt embeddings. 
The visual prompts can be clicked points, boxes, masks and scribbles, \etc. For the $i$-th target, we use $X_i^*\in R^{l_i\times 3}$ and $ y_i^* \in R^{l_i}$ to represent its prompt-specified pixels and corresponding segment IDs, where $l_i$ is the total number of prompt-specified pixels. To convert the visual prompts into image embeddings, we adopt the Visual Sampler strategy proposed in SEEM \cite{seem}. It samples $l^*$ points from the prompt-specified pixels for each target, and extracts point features from the 3rd scale image embeddings $F_3$ as its visual prompt embeddings: 
\begin{align}
\setlength{\abovedisplayskip}{3pt}
\setlength{\belowdisplayskip}{3pt}
P_{i}^* = \text{VisualSampler}(X_i^*,\ F_3),
\end{align}
where the shape of $P_i^* $ is ${l^* \times C}$.

The language prompt can be category names, such as `person', and textual expressions, such as `a person is skateboarding'. Following \cite{devlin2018bert,clip}, we feed a category name or an expression into a tokenizer to get its string tokens $L_i^* \in R^{l^* \times C^t}$, which are input into the CLIP text encoder to obtain the text embeddings. 
We then introduce a single cross-attention layer to achieve language-image embeddings interaction, where the query is text embeddings, the keys and values are flattened multi-scale image embeddings $F \in R^{C\times (\sum_{s=1}^3 H_sW_s)}$. This process can be formulated as:
\begin{align}
    \setlength{\abovedisplayskip}{3pt}
    \setlength{\belowdisplayskip}{3pt}
    P_i^*  = \text{Lang2Img-CA}(\text{CLIPTextEnc}(L_i^*) \cdot W^{t2v},\ F), 
\end{align}
where the shape of text embeddings $P_i^*$ is $l^* \times C$, and $l^*$ is the length of string tokens. The matrix $W^{t2v} \in R^{C^t\times C}$ maps the text embddings of dimension $C^t$ to the visual space of dimension $C$.  Note that we freeze the weights of CLIP text encoder to take advantage of the strong open vocabulary capabilities of CLIP.

\subsection{Unified Video Mask Decoder} \label{sec:uni_mask_dec}
The unified video mask decoder aims to decode the masks for prompt-specified targets, which can be described as:
\begin{equation}
    M_i^t = \text{MaskDec}(\{F_s^t\}_{s=1}^{4},\ P_i^*,\ y_i^*), \forall i \in [1, N_p], t \in [1, V] \nonumber
\end{equation}
where $P_i^*$ and $ y_i^*$ are prompt features and associated segment IDs for the $i$-th target, and $M_i^t$ is its predicted masks in the $t$-th frame. $N_p$ and $V$ are the total numbers of provided targets and  frames in the video, respectively.
To achieve our goal, we adapt the mask decoder of Mask2Former \cite{cheng2021mask2former,cheng2021mask2former-video}, which was initially designed for generic segmentation tasks with a set of learnable queries, by introducing a side stream that takes the mean of prompt features as input queries. As illustrated in the right yellow area of Fig. \ref{fig:overview}, our proposed unified video mask decoder comprises four key components: target-wise prompt cross-attention layer, an image cross-attention layer, a separated self-attention layer and a feed-forward network (FFN, which is omitted in Fig. \ref{fig:overview}).

\textbf{Initial prompt query (P$\rightarrow$Q).}
The visual/text prompt embeddings of the $i$-th target $P_i^* \in  R^{l^* \times C}$ consist of $l^*$ prompt tokens, which are point features from the visual prompt or string tokens of category names and expressions. We compute the average of all prompt tokens associated with it as the initial query for the target: $q_i^* = \sum\nolimits_{l=1}^{l^*}{P_{i, l}^*}$. If the input contains a video clip with $T$ frames, the initial query will be repeated $T$ times to generate a clip-level initial query $q_i^* \in R ^{T \times C}$.
Using the mean of prompt features as the initial query provides an informative and stable starting point for the unified video mask decoder.

\textbf{Prompt cross-attention (ProCA).} 
The initial query may not be sufficient to provide a distinct representation for targets, particularly for those with similar characteristics, such as `person' and `black T-shirt' in Fig. \ref{fig:overview}. To enhance the uniqueness in representation, we introduce an entity-wise prompt cross-attention layer to learn  prompt information to better differentiate between targets:
{\small
\begin{align}
    \setlength{\abovedisplayskip}{3pt}
    \setlength{\belowdisplayskip}{3pt}
    \text{ProCA}(q_i^*, P_i^*) = \text{Softmax}(\frac{q_i^* W^Q\ (P_i^* W^K)^T}{\sqrt{d_k}})\ P_i^* W^V,
\end{align}
}%
where the query is $q_i^*$, and the keys and values are the prompt tokens $P_i^*$. $W^Q, W^K$ and $W^V$ represent the projection weights. The ProCA layer is placed in the front of image cross-attention layer to avoid forgetting prompt information as the decoder layer goes deep.

\textbf{Image cross-attention and separated self-attention.} 
The ProCA layer facilitates the incorporation of prompt information, while the image cross-attention layer focuses on extracting entity details from the input frames. We only compute the image cross-attention between each frame's query and the corresponding image features to reduce the memory overhead. Furthermore, the separated self-attention (Sep-SA) layer serves for two purposes. On one hand, it isolates the interactions between learnable queries and prompt queries, minimizing unnecessary negative impacts. On the other hand, by flattening learnable/prompt queries in the time dimension, it facilitates content interactions of the target of interest across spatial and time domains. The Sep-SA layer can be formulated as:
\begin{equation}
    \text{SepSA}(\mathbf{q}, \mathbf{q}^*) = \text{SA}(\mathbf{q}, \mathbf{q})\  \& \ \text{SA}(\mathbf{q}^*, \mathbf{q}^*),
\end{equation}
where $\mathbf{q} \in R^{NT \times C}$ and $\mathbf{q}^* \in R^{N_pT \times C}$ represent flattened learnable and flattened prompt queries, respectively, and $N$ and $N_p$ are the numbers of them.

\textbf{Overall architecture.} In addition to the ProCA, image cross-attention and SepSA layers, the FFN further allows the mask decoder to learn non-linear relationships from data. These four key components constitute a transformer layer, and our unified video mask decoder is composed of nine such transformer layers.
In addition, there are two mask decoding streams, which share the same set of weights, to decode learnable queries and prompt queries, respectively. Note that the ProCA layer is omitted for learnable queries to simplify the representation. 

To obtain the predicted masks for the $t$-th frame, we linearly combine mask coefficients with the finest-scale feature map $F_4^t \in R^{C\times H/4\times W/4}$, \ie, $[{M}^t, {M}^{*t}] = f_\text{mask}([\mathbf{q}^t, \mathbf{q}^{*t}]) \cdot F_4^t$, where mask coefficients are generated by passing the output queries through a multi-layer perception, denoted as $f_\text{mask}$.

\section{Training and Inference} 
\subsection{Training Stages}\label{sec:train_settings}
Training Losses consist of three terms: pixel-wise mask supervision loss, classification loss, and ReID loss.
The training process of UniVS consists of three stages: image-level training, video-level training and long video fine-tuning. In the first stage, UniVS is trained on multiple image segmentation datasets, pretraining the model with image-level annotations for a good visual representation.
In the second stage, we feed a short clip of three frames to the pretrained model, and fine-tune it on video segmentation datasets to perceive entity changes over a short period of time. In the third stage, we employ long video sequences of more than five frames to further fine-tune the unified video mask decoder, encouraging it to learn more discriminative features and trajectory information over a longer time period. To optimize memory usage, we freeze the backbone weights in the last two stages and further freeze the pixel decoder in the final stage.  In each iteration, all samples within a batch come from the same dataset. We found that this sampling strategy can make the training more stable compared with the mixed sampling from different datasets. 

More detailed information about the training losses and stages can be found in the \textbf{Supplementary Material}.

\subsection{Unified Streaming Inference Process}\label{sec:uni_reference}

In UniVS, the model input can be a single frame or a clip of multiple frames. In this subsection, we take a single frame as input to elucidate the unified inference process for generic category-specified and prompt-specified VS tasks.

\begin{figure}[t]
     \vspace{-1mm}
     \centering
     \includegraphics[width=1.\linewidth]{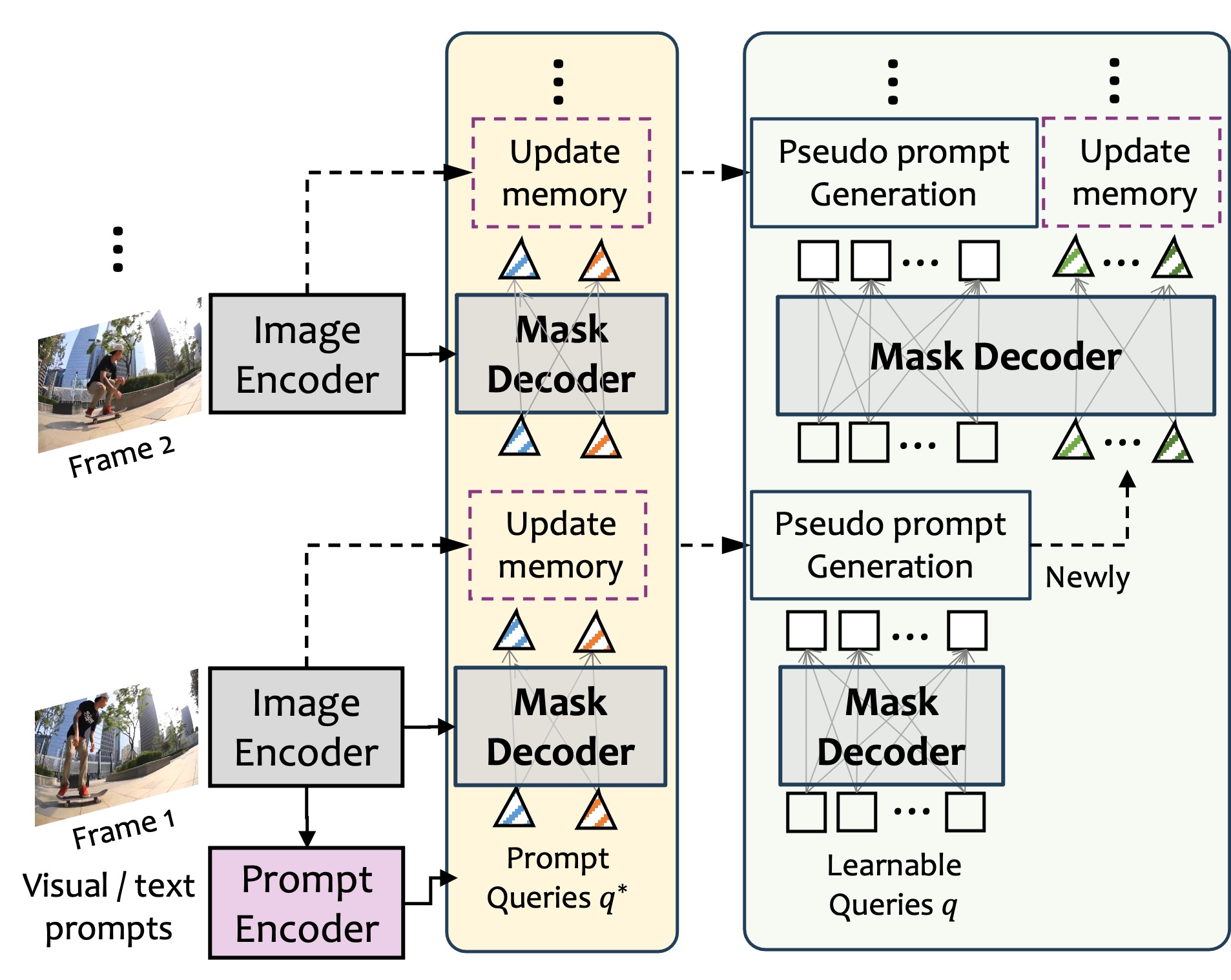}
     \vspace{-6mm}
     \caption{Inference process of our UniVS on prompt-specified and category-specified video segmentation tasks, respectively.}\label{fig:overview_inf}
     \vspace{-2mm}
\end{figure}

For \textbf{prompt-specified VS tasks}, UniVS takes video frames and visual/text prompts as input, and the inference process is illustrated in the yellow boxes of Fig. \ref{fig:overview_inf}. UniVS can process multiple targets simultaneously. First, the image encoder transforms the first frame into multi-scale image embeddings. Consequently, the prompt encoder converts visual/text prompts of the target into prompt tokens. In our design, each target has its dedicated memory pool to store associated prompt tokens, and its prompt query is obtained by averaging the tokens in the memory pool. These queries are used by the mask decoder to predict the masks of targets in the current frame, which are then used as visual prompts of targets and fed back to the prompt encoder, thereby updating the target's memory pool with new prompt information.
In short, UniVS utilizes the prompt information of target objects stored in the memory pool to identify and segment the targets in subsequent frames, eliminating the cumbersome post-matching step in other unified models like SEEM \cite{seem} and UNINEXT \cite{uninext}, where targets need to be filtered out from all predicted entities. 

For \textbf{category-specified VS tasks}, UniVS adopts a periodic object detection strategy and transforms the segmentation into a prompt-guided target segmentation problem. The detailed process is depicted in the light green box in Fig. \ref{fig:overview_inf}. First, UniVS employs learnable queries to identify all entity masks presented in the first frame, then employs non-maximum suppression (NMS) and classification thresholding to filter out redundant masks and those with low classification confidence. The remaining target objects also serve as their visual prompts, with which UniVS employs the prompt-guided target segmentation stream to directly predict their masks in the following frames, eliminating the need of cross-frame entity matching in previous methods. 
Furthermore, to identify newly appearing objects in subsequent frames, UniVS performs target detection for every a few frames using a learnable query and compares them to previously detected objects stored in a memory pool. We use the bi-Softmax approach \cite{pang2021quasi} to distinguish between old and new objects in the video.

\textbf{Remarks.} Existing VS methods mostly assume smooth object motion within a short clip to associate entities across frames. However, for videos containing complex trajectories or large scene changes, this assumption does not hold, resulting in a decline of tracking accuracy. In contrast, our proposed UniVS overcomes this limitation by using prompts as queries to achieve explicit mask decoding. 

\section{Experiments}
\subsection{Experimental Settings}
\textbf{Datasets.} 
The VIS datasets include YouTube-VIS 2019/2021 (YT19/21) \cite{yang2019video} and OVIS \cite{qi2021occluded}. The VSS and VPS datasets respectively include VSPW \cite{vspw} and VIPSeg. The VOS datasets include DAVIS \cite{davis}, YouTube-VOS 2018 (YT18) \cite{youtubevos}, MOSE \cite{mose} and BURST \cite{burst}.
RefVOS datasets include RefDAVIS \cite{refvos} and Ref-YouTube-VOS (RefYT) \cite{refvos}, while PVOS uses VIPOSeg \cite{xu2023viposeg} as dataset. The 2017 version of DAVIS and RefDAVIS are used  \cite{davis}. Note that VSPW, VIPSeg and VIPOseg share the same original videos but have different protocols and guidelines for annotation. Due to the high cost of annotating datasets, the video scenes contained in each dataset lack diversity, being mostly sports scenes or animal scenes. 

\textbf{Implementation Details.} We use Detectron2 \cite{wu2019detectron2} and follow Mask2Former \cite{cheng2021mask2former} baseline settings. For data augmentation, we use the large-scale jittering (LSJ) augmentation \cite{du2021lsj_aug} with a random scale sampled from range 0.5 to 4.0, followed by fixed-size cropping and padding to $1024 \times 1024$. We use the distributed training framework with 16 V100/A100 GPUs. Each mini-batch has 3 images or 1 video clip (2-7 frames) per GPU. ResNet50 \cite{he2016resnet} and SwinT/B/L \cite{swin} are adopted as the backbone networks, while we use the CLIP Text Encoder, whose corresponding visual encoder is ResNet50x4. We sample 32 points per visual prompt for visual prompt-guided targets. Considering the difference in the number of objects on various videos, we increase the number of learnable queries from 100 to 200 in the unified video mask decoder.

More detailed information about experimental settings can be found in the \textbf{Supplementary Material}.

\begin{table*}[!t]
\centering
\begin{subtable}{0.498\textwidth}
\footnotesize
\setlength{\tabcolsep}{0.5mm}{
    \linespread{2}
    \begin{tabular}{p{0.22\linewidth}p{0.115\linewidth}<{\centering}|p{0.085\linewidth}<{\centering}p{0.07\linewidth}<{\centering}p{0.09\linewidth}<{\centering}|p{0.08\linewidth}<{\centering}p{0.085\linewidth}<{\centering}|p{0.07\linewidth}<{\centering}p{0.07\linewidth}<{\centering}}
    \Xhline{0.8pt}
    \rowcolor{isabelline}
    \multicolumn{2}{c|}{Video Tasks} & \multicolumn{3}{c|}{VIS} &  \multicolumn{2}{c|}{VSS} &  \multicolumn{2}{c}{VPS} \\
    \hline
     \multirow{2}{*}{\!Method}& Back- & {\!YT19} & { YT21}  & { OVIS} &  \multicolumn{2}{c|}{ VSPW} & \multicolumn{2}{c}{ VIPSeg} \ \\
      & bone & mAP & mAP & mAP & mIoU & \!mVC$_8$ & \!VPQ & \!STQ  \\
    \hline
    \!IDOL\cite{IDOL}         & R50 & 49.5 & 43.9  & 30.2 & - & - & - & -  \\
    \!MinVIS\cite{huang2022minvis} & R50 & 47.4 & 44.2  & 25.0 & -  & - & - & -  \\
    \!MDQE\cite{li2023mdqe}     & R50 & *    & 44.5  & 29.2 & - & - & - & -  \\
    \!GenVIS-off\cite{GenVIS}   & R50 & \textbf{51.3} & \textbf{46.3}  & \textbf{34.5} & - & - & - & -  \\
    \!TCB\cite{vspw}            & R101 & - & - & - & \textbf{37.8} & \textbf{87.8}  & - & -  \\
    \!ClipPanoFCN\cite{vipseg}  & R50 & - & - & - & - & - & \textbf{22.9} & \textbf{31.5}  \\
    \hline
    \hline
    \!IDOL\cite{IDOL}         &  SwinL & 49.5 & 43.9 & 42.6 & - & - & - & - \\
    \!\!MinVIS\cite{huang2022minvis}  &  SwinL & -    & 55.3 & 41.6 & - & - & - & - \\ 
    \!GenVIS-off\cite{GenVIS} &  SwinL & \textbf{63.8} & \textbf{60.1} & \textbf{45.4} & - & - & - & - \\ 
    \!CFFM\cite{CFFM_vss}     & MiTB5 & - & - & - & \textbf{49.3} & \textbf{90.8}  & - & - \\
    \Xhline{0.8pt}
    \end{tabular}
    }
    \caption{\small Category-specified VS models
} \label{tab:sota_r50_category}
\end{subtable}
\begin{subtable}{0.495\textwidth}
\footnotesize
\setlength{\tabcolsep}{0.5mm}{
    \linespread{2}
    \begin{tabular}{p{0.22\linewidth}p{0.115\linewidth}<{\centering}|p{0.09\linewidth}<{\centering}p{0.09\linewidth}<{\centering}|p{0.13\linewidth}<{\centering}p{0.11\linewidth}<{\centering}|p{0.14\linewidth}<{\centering}}
    \Xhline{0.8pt}
    \rowcolor{isabelline}
    \multicolumn{2}{c|}{Video Tasks} & \multicolumn{2}{c|}{VOS} & \multicolumn{2}{c|}{RefVOS} & PVOS \\
    \hline
     \multirow{2}{*}{\!Method}& Back- & {DAVIS} &{ YT18\ } & \!RefDAVIS & \!RefYT & \!\!VIPOSeg \ \\
      &bone & $G^{th}$ & $G^{th}$ & J\&F & J\&F & $G^{th\&sf}$ \\
    \hline
    \!XMem\cite{xmem}            & R50 & \textbf{86.2} & 85.7  & - & - & - \\ 
    \!DeAOT\cite{yang2022deaot}  & R50 & 85.2 & \textbf{86.0} & - & - & - \\ 
    \!ReferFormer\cite{referformer}  & R50 & - & -  & 58.5 & 55.6 & -\\
    \!OnlineRefer\cite{wu2023onlinerefer}  & R50 & - & -  & 59.3 & 57.3 & -\\
    \!SgMg\cite{Miao_2023_ICCV_sgmg}   & SwinT & - & -  & \textbf{61.9} & \textbf{62.0} & -\\
    \!PAOT\cite{xu2023viposeg}         & R50 & - & -  & - & - & \textbf{75.4} \\
    \hline
    \hline
    \!DeAOT\cite{yang2022deaot} & Swin-B & \textbf{86.2} & \textbf{86.2}  & - & - & - \\ 
    \!OnlineRefer\cite{wu2023onlinerefer}  & Swin-L & - & -  & \textbf{64.8} & 63.0 & - \\ 
    \!SgMg\cite{Miao_2023_ICCV_sgmg}  & SwinB & - & -  & 63.3 & \textbf{65.7} & -\\  
    \!PAOT\cite{xu2023viposeg}        & SwinB  & - & -  & - & - &  \textbf{75.3} \\  
    \Xhline{0.8pt}
    \end{tabular}
    }
\caption{\small Prompt-specified VS models
} \label{tab:sota_r50_prompt}
\end{subtable}
\vspace{-7mm}
\caption{\small Quantitative performance comparison on individual VS models, which are specifically designed based on the characteristics of each task. `-' means the model lacks this capability. The best results are shown in \textbf{bold}.
} \label{tab:sota_r50_individual}
\vspace{-2mm}
\end{table*}
\begin{table*}[!t]
\centering
{\footnotesize
\begin{tabular}{p{0.11\linewidth}p{0.075\linewidth}<{\centering}p{0.035\linewidth}<{\centering}p{0.048\linewidth}<{\centering}|p{0.025\linewidth}<{\centering}p{0.023\linewidth}<{\centering}p{0.03\linewidth}<{\centering}|p{0.02\linewidth}<{\centering}p{0.028\linewidth}<{\centering}|p{0.022\linewidth}<{\centering}p{0.022\linewidth}<{\centering}|p{0.025\linewidth}<{\centering}p{0.035\linewidth}<{\centering}|p{0.032\linewidth}<{\centering}p{0.032\linewidth}<{\centering}|p{0.05\linewidth}<{\centering}}
\Xhline{0.8pt}
\rowcolor{isabelline}
\multicolumn{4}{c|}{Video Tasks} & \multicolumn{3}{c|}{VIS} &  \multicolumn{2}{c|}{VSS} &  \multicolumn{2}{c|}{VPS} & \multicolumn{2}{c|}{VOS} & \multicolumn{2}{c|}{RefVOS} & PVOS \ \\
\hline
 \multirow{2}{*}{\!\!Method} & \multirow{2}{*}{Backbone} & \!\!Joint & \multirow{2}{*}{ \!\!\!\!Universal} & { \!YT19} & { YT21}  & { \!OVIS} &  \multicolumn{2}{c|}{ VSPW} & \multicolumn{2}{c|}{ VIPSeg} & \!\!DAVIS & { \!YT18 }   & \!\!DAVIS & \!\!\!RefYT & \!\!\!VIPOSeg\ \\
  & & \!\!\!\!\!Training & & mAP & mAP &  mAP & \!\!mIoU & \!\!\!mVC$_8$ & \!VPQ & \!STQ & $G^{th}$ & $G^{th}$ & J\&F & J\&F &$G^{th\&sf}$ \\
\hline 
\!\!VideoK-Net\cite{videoknet}  & ResNet50 & \crossmark & \crossmark & 40.5  &  * & * & * & * & 26.1 & 33.1 & - & - & - & - & - \\ 
\!\!Video-kMax\cite{videokmax}  & ResNet50& \crossmark & \crossmark & * & * & * &  {44.3} & 86.0 & 38.2 & 39.9 & - & - & - & - & - \\ 
\!\!Tube-Link\cite{tubelink}   & ResNet50 & \crossmark & \crossmark &  {52.8} &  {47.9} & 29.5 & 42.3 &  {86.8} & 39.2 & 39.5 & - & - & - & - & - \\ 
\!\!DVIS-off\cite{dvis}     & ResNet50 & \crossmark & \crossmark & 52.6 & 47.4 &  {33.8} & * & * & \textbf{43.2} & 42.8 & - & - & - & - & - \\ 
\!\!UniRef\cite{Wu_2023_ICCV_uniref}   & ResNet50    & $\checkmark$ & \crossmark & - & - & - & - & - & - & - & * & \textbf{81.4} & 63.5 & 60.6 & * \\  
\!\!TarVIS\cite{tarvis}    & ResNet50   & $\checkmark$ & \crossmark &  * & \textbf{48.3} & 31.1 & * & * &  33.5 & 43.1 & \textbf{82.6} & * & - & - & - \\  
\!\!UNINEXT\cite{uninext}    & ResNet50  & $\checkmark$ & \crossmark & \textbf{53.0} & *  & \textbf{34.0} & - & -  & - & - & 74.5 & 77.0 & \textbf{63.9} & \textbf{61.2} & - \\ 
\!\!UniVS     & ResNet50  & $\checkmark$ & $\checkmark$   & 47.4 & 46.6  & 30.8 &  \textbf{48.2} & \textbf{88.5} & 38.6  & \textbf{45.8} & 70.5 & 69.2 & 57.9 & 56.2 & \textbf{60.2} \\ 
\!\!\textcolor{gray}{UniVS} & \textcolor{gray}{SwinT} & \textcolor{gray}{$\checkmark$} & \textcolor{gray}{$\checkmark$}   & \textcolor{gray}{52.4} & \textcolor{gray}{51.6}  & \textcolor{gray}{33.0} & \textcolor{gray}{51.3} & \textcolor{gray}{89.4} & \textcolor{gray}{38.9} & \textcolor{gray}{51.7} & \textcolor{gray}{71.7} & \textcolor{gray}{70.3} & \textcolor{gray}{58.5} & \textcolor{gray}{56.2} & \textcolor{gray}{62.3} \\ 
\hline
\hline
\!\!TubeFormer\cite{tubeformer}   &  A-R50$\times$64 & \crossmark & \crossmark & 47.5 & 41.2 & * & 63.2 & 92.1 & *  & * & - & - & - & -& - \\
\!\!VideoK-Net+\cite{videoknet} & SwinB & \crossmark & \crossmark & 51.4 & * & * & * & * & 39.8 & 46.3 & - & - & - & - & - \\
\!\!Video-kMax\cite{videokmax}   & ConvNeXtL & \crossmark & \crossmark & * & * & * & \textbf{63.6} & 91.8 & 51.9 & 51.7 &  - & - & -  & -& - \\
\!\!Tube-Link\cite{tubelink}    & SwinL & \crossmark & \crossmark & 64.6 & 58.4 & * & 59.7 & 90.3 & * & * & - & - & - & - & - \\  
\!\!DVIS-off\cite{dvis}     & SwinL & \crossmark & \crossmark & \textbf{64.9} & 60.1 & \textbf{49.9} & * & * & \textbf{57.6} & 55.3  & - & - & - & -& - \\ 
\!\!UniRef\cite{Wu_2023_ICCV_uniref}       & SwinL   & $\checkmark$ & \crossmark & - & - & - & - & - & - & - & * & \textbf{82.6} &  66.3 & \textbf{67.4} & * \\  
\!\!TarVIS\cite{tarvis}       & SwinL    & $\checkmark$ & \crossmark &  *   & \textbf{60.2} & 43.2 & * & * & 48.0 & 52.9 & \textbf{85.2} & * & - & - & -  \\
\!\!UNINEXT\cite{uninext}      & ConvNeXtL & $\checkmark$ & \crossmark & 64.3 & *    & 41.1 & - & - & - & - & 77.2 & 78.1 & \textbf{66.7} & 66.2 & -\\ 
\!\!\textcolor{gray}{UNINEXT}\cite{uninext}      & \textcolor{gray}{ViT-H}    & \textcolor{gray}{ $\checkmark$} & \textcolor{gray}{\crossmark} & \textcolor{gray}{66.9} & \textcolor{gray}{ *}    & \textcolor{gray}{49.0} & \textcolor{gray}{-} & \textcolor{gray}{-} & \textcolor{gray}{-} & \textcolor{gray}{-} & \textcolor{gray}{81.8} &\textcolor{gray}{78.6} & \textcolor{gray}{72.5} & \textcolor{gray}{70.1} & \textcolor{gray}{-}\\ 
\!\!UniVS        & SwinB   & $\checkmark$ & $\checkmark$ & 57.8 & 56.5 & 39.0 & 59.4 & 90.4 & 46.7 & 56.1  & 75.0 & 70.9 & 58.6 & 57.4 & 68.2 \\ 
\!\!UniVS        & SwinL   & $\checkmark$ & $\checkmark$ & 60.0 & 57.9 & 41.7 & 59.8  & \textbf{92.3}  & 49.3 & \textbf{58.2}  & 76.2 & 71.5 & 59.4 & 58.0 & \textbf{68.6} \\  
\Xhline{0.8pt}
\end{tabular}
}
\vspace{-3mm}
\caption{\small Overall quantitative performance comparison of unified VS models. `-' means the model lacks this capability and `*' means the result is not reported. The results of UniVS with SwinT backbone and UNINEXT with ViT-H backbone are also listed in gray color. However, due to use of different backbones, they are not considered in the performance comparison. The best results are shown in \textbf{bold}.
} \label{tab:sota_strong_vid2}
\vspace{-2mm}
\end{table*}

\subsection{Video Benchmark Results}
We compare the results of recent high-performance VS models. The results of individual models trained on a single task/dataset are shown in Table \ref{tab:sota_r50_individual}, and the results of unified models trained individually or jointly on different tasks are shown in Table \ref{tab:sota_strong_vid2}. To assess the generalization capability of competing models, we present the quantitative performance comparison on 10 benchmarks of six VS tasks, including VIS, VSS, VPS, VOS, RefVOS and PVOS. The detailed quantitative comparison on more image/video benchmarks can be found in the \textbf{Supplementary Material}.

\textbf{Individual models} are specifically designed based on the characteristics of each task, resulting in high performance on each benchmark, as shown in Table \ref{tab:sota_r50_individual}. For VIS task, GenVIS-off \cite{GenVIS} with sequential learning achieves impressive performance of 51.3, 46.3 and 34.5 mAP on YT19, YT21 and OVIS, respectively. For the recently proposed VSS/VPS tasks, the baselines \cite{vspw,vps} only obtain decent performance.
For VOS task, XMem\cite{xmem} and DEAOT \cite{yang2022deaot} employ long-term information propagation modules, achieving state-of-the-art performance of 86.2 on DAVIS and 86.0 on YT18, respectively. For RefVOS task, SgMg \cite{Miao_2023_ICCV_sgmg} achieves the best performance of 62.0 on RefYT by using a spectrum-guided multi-granularity approach. For the newly proposed PVOS task, PAOT \cite{xu2023viposeg} extends the typical VOS method DEAOT, obtaining 75.4 on VIPOSeg. 
 
\textbf{Unified models trained individually} on different tasks include VideoK-Net \cite{videoknet}, Video-kMax \cite{videokmax}, Tube-Link \cite{tubelink}, DVIS \cite{dvis} and TubeFormer \cite{tubeformer}, as listed in Table \ref{tab:sota_strong_vid2}, which tend to handle both thing and stuff categories within a single framework.
Video-kMax with clip k-means mask transformer achieves high performance on VSS task, reaching 44.3 mIoU on VSPW. By using the strong Mask2Former\cite{cheng2021mask2former} as baseline, Tube-Link and DVIS inherit its ability to handle both thing and stuff classes, achieving remarkable performance on VIS and VPS tasks, as evidenced by the 33.8 mAP on OVIS and 43.2 VPQ on VIPSeg. It should be noted that the above unified models are trained individually on each dataset so that they can achieve excellent performance on each benchmark; however, this makes them lack the generalization ability to other datasets. Additionally, they cannot handle prompt-specified VS tasks, including VOS, RefVOS and PVOS.

\textbf{Unified models trained jointly} on different tasks aim to accommodate as many VS tasks as possible within a single model, where all tasks can be accomplished using the same set of trained weights. As shown in the top rows of Table \ref{tab:sota_strong_vid2}, UniRef \cite{Wu_2023_ICCV_uniref} unifies all prompt-specified VS tasks, but it fails to deal with category-specified VS tasks, such as VIS, VSS and VPS. TarVIS \cite{tarvis} can handle most VS tasks except for RefVOS due to the lack of language encoding capabilities. UNINEXT \cite{uninext} focuses solely on object-centric segmentation without considering entity segmentation of stuff categories, thereby being incapable of handling VSS, VPS and PVOS tasks. In contrast, our UniVS is the only approach that accommodates all VS tasks within a single model, demonstrating the highest generalization capability in universal segmentation.

Specifically, UniRef achieves outstanding performance on VOS and RefVOS tasks, reaching 81.4 on YT18. TarVIS achieves state-of-the-art performance on datasets with simple scenes, such as 48.3 mAP on YT21 of VIS task and 82.6 on DAVIS of VOS task. UNINEXT utilizes a large amount of image/video data to jointly train the model, allowing it to achieve state-of-the-art performance in instance-level segmentation benchmarks, such as 53.0/34.0 mAP on YT19/OVIS for VIS task and 63.9/61.2 on DAVIS and RefYT for RefVOS task. Our UniVS achieves comparable performance on VIS task, top-ranked performance on VSS/VPS tasks and slightly lower performance in prompt-specified VS tasks. For VSS and VPS tasks, UniVS brings 2$\sim$4\% performance improvement, reaching 48.2 mIoU and 88.5 mVC$_8$ on VSPW and 45.8 STQ on VIPSeg.

Compared with UniRef, TarVIS and UNINEXT, our UniVS can handle category-specified VS tasks, text prompt-guided segmentation and entity segmentation for stuff categories simultaneously. Pursuing a more universal segmentation capability, however, may sacrifice performance in individual tasks due to task conflicts and limited video data. 
Compared to UniRef, which utilizes heavy temporal propagation modules to pass reference frames and masks to the current frame, UniVS adopts a parameter-free Visual Sampler to extract prompt information, resulting in a slight decrease in performance.
Compared to UNINEXT, which uses 900 object queries, we train UniVS  using only 200 object queries with fewer video data. Additionally, UniVS needs to accommodate stuff categories in object segmentation. These two factors explain the slightly lower performance of our approach in instance-level segmentation benchmarks compared to UNINEXT. In future work, more diverse training video data and long-term information propagation modules will be explored to improve the performance of our unified segmentation models. 

Our results with stronger backbones further validate the aforementioned conclusion, as shown in the bottom rows of Tables \ref{tab:sota_r50_individual} and \ref{tab:sota_strong_vid2}. Our UniVS is the only model that is capable of handling all the six VS tasks within a single framework, setting new state-of-the-art performance on VSS and VPS tasks, and achieving 92.3 mVC$_8$ on VSPW and 58.2 STQ on VIPSeg. Overall, our UniVS can obtain an appropriate balance between performance and universality capability.

\begin{table}[!t]
\centering
\begin{subtable}{0.495\textwidth}
\small
\setlength{\tabcolsep}{0.5mm}{
    \linespread{2}
    \begin{tabular}{p{0.1\linewidth}<{\centering}p{0.12\linewidth}<{\centering}|p{0.11\linewidth}<{\centering}p{0.11\linewidth}<{\centering}|p{0.1\linewidth}<{\centering}p{0.1\linewidth}<{\centering}|p{0.11\linewidth}<{\centering}|p{0.13\linewidth}<{\centering}}
    \Xhline{0.8pt}
    \rowcolor{isabelline}
    \multicolumn{2}{c|}{Video Tasks} & \multicolumn{2}{c|}{VIS} &  \multicolumn{2}{c|}{VPS} & \multicolumn{1}{c|}{VOS} & RefVOS \\
    \hline
    \multirow{2}{*}{P$\rightarrow$Q} & \multirow{2}{*}{ProCA}  & { YT21}  & {OVIS} & \multicolumn{2}{c|}{ VIPSeg} &{ YT18\ }  & RefYT \ \\
           & & mAP & mAP & VPQ & STQ & $G^{th}$  & J\&F\\
    \hline
    \crossmark   & \crossmark   & 45.9 & 17.2 & 40.1 & 37.9 & - & - \\
    $\checkmark$ & \crossmark   & 39.9 & 10.0 & 40.6 & 37.3 & 58.6 & 42.1 \\ 
    $\checkmark$ & $\checkmark$ & 52.7 & 21.7 & 35.4 & 49.2 & 67.4 & 54.9 \\
    \Xhline{0.8pt}
    \end{tabular}
    }
\vspace{+0.5mm}
\caption{\small Prompt as queries (P$\rightarrow$Q) and prompt cross-attention (ProCA)
} \label{tab:abl_prompt}
\vspace{+1mm}
\end{subtable}
\begin{subtable}{0.495\textwidth}
\small
\setlength{\tabcolsep}{0.5mm}{
    \linespread{2}
    \begin{tabular}{p{0.075\linewidth}<{\centering}p{0.145\linewidth}<{\centering}|p{0.11\linewidth}<{\centering}p{0.11\linewidth}<{\centering}|p{0.1\linewidth}<{\centering}p{0.1\linewidth}<{\centering}|p{0.11\linewidth}<{\centering}|p{0.13\linewidth}<{\centering}}
    \Xhline{0.8pt}
    \rowcolor{isabelline}
    \multicolumn{2}{c|}{Video Tasks} & \multicolumn{2}{c|}{VIS} &  \multicolumn{2}{c|}{VPS} & \multicolumn{1}{c|}{VOS} & \!RefVOS \\
    \hline
    \multirow{2}{*}{\footnotesize \!Stage 3} & \multirow{2}{*}{ \footnotesize Tracker} & {\!YT21}  & {\!OVIS} & \multicolumn{2}{c|}{ VIPSeg} & YT18  & RefYT \ \\
         & & mAP & mAP & \!VPQ & \!STQ & $G^{th}$ & J\&F \\
    \hline
    \crossmark & { \footnotesize Similarity}  & 50.7 & 15.0 & 35.9 & 44.5 & 60.8 & - \\
    \crossmark & VS$\rightarrow$P   & 52.7 & 21.7 & 35.4 & 49.2 & 67.4 & 54.9 \\
    $\checkmark$ & VS$\rightarrow$P   & 54.7 & 23.6 & 38.6 & 45.8 & 69.2 & 56.2 \\ 
    \Xhline{0.8pt}
    \end{tabular}
    }
\vspace{+0.5mm}
\caption{\small Unified training and inference, where VS$\rightarrow$P refers to the transformation from all VS tasks to prompt-guided target segmentation.
} \label{tab:abl_reference}
\vspace{-1mm}
\end{subtable}
\caption{\small Ablation studies on (a) prompt as queries and (b) unified training and inference on VS tasks. `-' means that the model lacks this capability.
For VIS task, the results are evaluated in the development set (1/10 of the training set, excluded during training).
}
\vspace{-1mm}
\end{table} 
\subsection{Ablation Studies}
We analyze UniVS through a series of ablation studies using the ResNet50 backbone \cite{he2016resnet}. To test the generality of the proposed components for universal video segmentation, ablation studies are performed on various VS tasks.

\textbf{Prompts as Queries and ProCA.} We validate the importance of each component by adding them one at a time. As shown in Table \ref{tab:abl_prompt}, by introducing the prompt as queries (P$\rightarrow$Q), the baseline can handle prompt-specific VS tasks but achieve relatively lower performance on all VS tasks.  Moreover, by introducing the prompt cross-attention (ProCA) to further extract comprehensive prompt features, the performance on most VS tasks is significantly improved, especially on OVIS for VIS task and YT18 for VOS task, whose performance is increased by $\sim$10\%.

\begin{figure}[!tbh]
     \centering
     \includegraphics[width=0.98\linewidth]{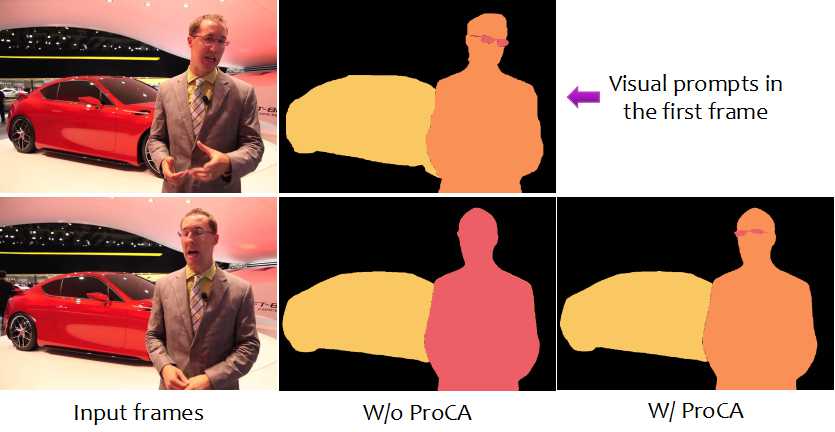}
     \vspace{-3mm}
     \caption{\footnotesize Qualitative results of UniVS w/o and w/ ProCA on VOS task.}\label{fig:proca}
\end{figure}
The predicted masks of UniVS without and with the ProCA layer are shown in Fig. \ref{fig:proca}. It is evident that, for the uncommon object `glasses', the ProCA layer could provide precise prompt features to prevent the mask from spilling over to other objects like `person'. 

\textbf{Unified Training and Inference.} 
In Table \ref{tab:abl_reference}, we study the performance when using different ways to associate entities across video clips. Using query embedding similarity as a tracker fails to differentiate entity trajectories in complex scenarios, leading to poor performance. 
By adopting our unified streaming inference process in Sec. \ref{sec:uni_reference}, which transfers all VS tasks to the prompt-guided target segmentation, the performance of all VS tasks improves significantly, particularly in complex scenes (with an increase of $\sim$6\%). Additionally, introducing the third training stage improves the model's ability to capture medium-term object motion trajectories, bringing $\sim$2\% performance improvement.
\vspace{+2mm}
\section{Conclusion}
In this paper, we attempted to accommodate all video segmentation tasks within a single model, and proposed a novel unified architecture, namely UniVS, by using prompts as queries. We averaged the prompt features stored in the memory pool as the initial query of the prompt-guided target, and introduced a target-wise prompt cross-attention layer to integrate comprehensive prompt features, and converted different VS tasks into prompt-guided target segmentation during inference.
Extensive experimental results demonstrated that UniVS achieved competitive or even better performance on category-specified VS tasks, while achieved slightly lower performance on prompt-specified VS tasks. Overall, by using a single model with the same set of trained model parameters, UniVS resulted in a commendable balance between performance and universality. Its potentials can be further released with more training video-language data.   

\newpage
{\small
\bibliographystyle{ieee_fullname}
\bibliography{main}
}

\newpage

\setcounter{section}{0}
\addcontentsline{toc}{section}{Appendices}
\renewcommand{\thesection}{\Alph{section}}

In this supplementary file, we provide the following materials:
\begin{itemize}[leftmargin=0.25cm, itemindent=1cm]
    \setlength{\itemsep}{2pt}
    \setlength{\parskip}{0pt}
    \setlength{\parsep}{0pt}
    \item[A.] {Datasets and Evaluation Metrics}
    \item[B.] {Training and Inference Details}
    \item[C.] {More Ablation Studies}
    \item[D.] {More Visualization Results}
\end{itemize}

\section{Datasets and Evaluation Metrics}
Video segmentation (VS) tasks can be divided into two groups: category-specified and prompt-specified VS tasks. Table \ref{tab:vs_data} summarizes the statistics of different VS datasets.
\begin{table*}[h]
\centering
\footnotesize
\setlength{\tabcolsep}{1.5mm}
    \begin{tabular}{c|ccc|c|c|c|cccc|cc}
    \Xhline{0.8pt}
     Tasks & \multicolumn{3}{c|}{VIS} & VSS & VPS & PVOS & \multicolumn{4}{c|}{VOS} & \multicolumn{2}{c}{RefVOS} \\
     Datsets & \cellcolor{LightYellow} \!\!YT19 & \cellcolor{LightYellow}\!\!YT21 & \cellcolor{lightmauve} OVIS & \cellcolor{LightCyan} VSPW & \cellcolor{LightCyan} VIPSeg  & \cellcolor{LightCyan} \!\!VIPOSeg & \cellcolor{cottoncandy}\!DAVIS & \cellcolor{LightYellow} \!YT18 & \cellcolor{lightmauve} \!\!MOSE & \!\!BURST & \cellcolor{cottoncandy}\!RefDAVIS & \cellcolor{LightYellow} RefYT \\
    \hline
    Videos   & 2.8k & 3.8k  & 1.0k & 3.5k & 3.5k & 3.5k  & 0.09k  & 4.4k  & 2.1k & 1.9k  & 0.09k & 4.0k \\ 
    Images   & 97k  & 92k  & 51k  & 252k & 85k  & 85k &  6k       & 97k           & 96k   & 196k  & 6k   & 97k \\  
    Masks    & 131k & 232k & 296k &  -   & 926k & 926k  &  13k      & 197k          & 431k  & 600k    & 13k  & 197k  \\
    Classes  &  40 &  40 & 25 & 124      & 124  & 124  & -         & 94            & 36    & 482  & -    & - \\ 
    Expressions & - & - & - & - & - & - & - & - & - & - & 1.5k & 28k \\
    Thing    & $\checkmark$ & $\checkmark$ & $\checkmark$ & $\checkmark$ & $\checkmark$  & $\checkmark$  & $\checkmark$ & $\checkmark$ & $\checkmark$ & $\checkmark$ & $\checkmark$ & $\checkmark$ \\
    Stuff    & \crossmark   & \crossmark   & \crossmark   & $\checkmark$ & $\checkmark$ & $\checkmark$   & \crossmark   & \crossmark   & \crossmark   & \crossmark   & \crossmark & \crossmark \\
    Exhaustive & \crossmark & \crossmark   & \crossmark   & $\checkmark$ & $\checkmark$ & $\checkmark$ & \crossmark    & \crossmark    & \crossmark   & \crossmark    & \crossmark & \crossmark \\
    \Xhline{0.8pt}
    \end{tabular}
\vspace{-2mm}
\caption{ Statistics of different video segmentation datasets. The datasets labeled with the same color share the source video data but have different annotation formats, such as VSPW, VIPSeg and VIPOSeg.
} \label{tab:vs_data}
\end{table*}
\begin{table*}[h]
\centering
{\footnotesize 
\setlength{\tabcolsep}{1.35mm}
    \begin{tabular}{cc|cccc|ccccccc|cccc}
\Xhline{0.8pt}
\rowcolor{isabelline}
 & & \multicolumn{4}{c|}{Images} &  \multicolumn{7}{c|}{Videos}  &  & & & \\
\rowcolor{isabelline}
 \multicolumn{2}{c|}{\multirow{-2}{*}{Datasets}} & IS & PS & IS & Ref & VIS & VIS & VPS & VOS & VOS & VOS & Ref & \multicolumn{4}{c}{\multirow{-2}{*}{Settings}}\\
\hline
 Training & \!\!Frames & SA1B & \!\!COCO & LVIS & \!\!RefCOCO & \!YT21 & OVIS  & VIPSeg & YT18 & MOSE & BURST & \!\!RefYT & GPUs & Lr & \!\!Max Iter & Step \\
\hline
Stage 1 & 1 & 1.0  & 1.0  & 0.5  & 1.0  & -  & - &- & -  & - &- & - & 16 & 1e-4 & 354k & 342k \\ 
Stage 2 & 3 & 0.25 & 0.5 & 0.25 & 0.35 & 0.25 & 0.35 & 0.5 & 0.25 & 0.25 & 0.25 & 0.35  & 8 & 5e-5 & 708k & 684k \\ 
Stage 3 & 5-7 & 0.25 & 0.5 & 0.25 & 0.35 & 0.25 & 0.35 & 0.5 & 0.25 & 0.25 & 0.25 & 0.35 & 8 & 5e-5 & 177k & 162k \\ 
\Xhline{0.8pt}
\end{tabular}
}
\vspace{-2mm}
\caption{\small Implementation details in training. The sampling weights of each dataset during different training stages are given. `-' means that the dataset is not used. `Step' means the iterations when the learning rate is reduced.
} \label{tab:training_details}
\vspace{-2mm}
\end{table*}

\subsection{Category-specified VS Datasets}
 Category-specified VS tasks include video instance segmentation (VIS) \cite{yang2019video,qi2021occluded}, video semantic segmentation (VSS) \cite{vspw} and video panoptic segmentation (VPS) \cite{vps,vipseg}, where the object categories need to be specified.

\textbf{Video Instance Segmentation (VIS)} involves identifying and segmenting individual objects within each frame of a video while maintaining temporal consistency across frames. There are two large-scale VIS datasets: YouTube-VIS \cite{yang2019video} series and OVIS \cite{qi2021occluded}.
\textbf{YouTube-VIS} \cite{yang2019video} has three versions: YT19/21/22. The commonly used version is YT21, which contains 2,985 training, 421 validation, and 453 test videos over 40 `thing' categories. The number of frames per video is between 19 and 36. \textbf{OVIS} \cite{qi2021occluded} targets at distinguishing occluded objects in long-time videos (up to 292 frames), which includes 607 training, 140 validation, and 154 test videos, scoping 25 `thing' categories. 
VIS task adopts average precision (AP$_*$), average recall (AR$_*$) and the mean value of AP (mAP) as metrics for evaluation.

\textbf{Video Semantic Segmentation (VSS)} needs to perform pixel-level labeling of semantic categories in each frame of a video. 
\textbf{VSPW} \cite{vspw} is the first large-scale video scene parsing dataset, containing 3,536 annotated videos and 124 semantic thing/stuff classes. 
VSS uses mIoU, mVC$_8$ and  mVC$_{16}$ as metrics for evaluation, 
where mean video consistency (mVC$_*$) evaluates the category consistency among long-range adjacent frames (`*' indicates the number of frames in a video clip).

\textbf{Video Panoptic Segmentation (VPS)} combines VIS and VSS tasks by simultaneously identifying and tracking individual object instances while assigning semantic labels to each pixel. The goal is to achieve a comprehensive understanding of both instance-level and semantic-level information across the video sequence.
\textbf{VIPSeg} \cite{vipseg} is the first large-scale VPS dataset in the wild, which shares the original videos from the VSPW dataset. VIPSeg has pixel-level panoptic annotations, covering a wide range of real-world scenarios and categories. There are two commonly used evaluation metrics for the VPS task: VPQ \cite{vps} and STQ \cite{step_stq}. Video Panoptic Quality (VPQ) computes the average mask quality by performing tube IoU matching across a small span of frames. Segmentation and Tracking Quality (STQ) is proposed to measure the segmentation quality and long term tracking quality simultaneously.

\subsection{Prompt-specified VS Datasets}
Prompt-specified VS focuses on identifying and segmenting specific targets throughout the video, where visual prompts or textual descriptions of the targets need to be provided. It includes video object segmentation (VOS) \cite{Perazzi2016vos}, panoptic VOS (PVOS) \cite{xu2023viposeg} and referring VOS (RefVOS) \cite{refvos}.

\textbf{Video Object Segmentation (VOS)} segments a particular object throughout the entire video given only the object mask at the first frame, which can be viewed as the extension of interactive segmentation from spatial to temporal dimension. \textbf{DAVIS} \cite{davis}, an early proposed VOS dataset, contains a total of 90 videos. \textbf{YouTube-VOS} (YT18) \cite{youtubevos} consists of 4,453 short video clips with 94 different object categories. \textbf{MOSE} \cite{mose} targets at complex video object segmentation, whose videos partially inherit from OVIS \cite{qi2021occluded}. MOSE contains 2,149 video clips and 36 object categories.
To evaluate the performance, region jaccard $J$ and countour accuracy $F$ are computed for `seen' and `unseen' classes separately, denoted by subscripts $s$ and $u$. $G^{th}$ is the average  ($J\&F$) over both seen and unseen classes.

\textbf{Panoptic VOS (PVOS)} extends the above VOS task by taking stuff classes into account. Based on the VIPSeg dataset, \textbf{VIPOSeg} \cite{xu2023viposeg} is developed for PVOS. It contains exhaustive object annotations and covers various real-world object categories, which are carefully divided into subsets of thing/stuff and seen/unseen classes for comprehensive evaluation. This newly proposed benchmark uses eight separate metrics, including four mask IoUs for seen/unseen thing/stuff and four boundary IoUs \cite{cheng2021boundary_iou} for seen/unseen thing/stuff, respectively. The overall performance $G^{th\&sf}$ is the average of these eight metrics. 

\textbf{Referring VOS (RefVOS)} aims to segment the target object in a video based on the natural language description, which is a challenging multi-modal segmentation task. 
\textbf{RefDAVIS} and \textbf{RefYT} \cite{refvos} are two RefVOS datasets based on DAVIS and YouTube-VOS \cite{youtubevos}, respectively. RefYT is a large-scale benchmark covering 3,978 videos with around 28K language descriptions. The evaluation metrics include region similarity ($J$), contour accuracy ($F$) and their average value ($J\&F$). 

\section{Training and Inference Details}
\subsection{Training Losses} 
There are three terms in the training loss:
\begin{equation}
    L = \lambda_\text{mask} L_\text{mask} + \lambda_\text{cls} L_\text{cls} + \lambda_\text{reid} L_\text{reid},
\end{equation}
where $\lambda_\text{mask}, \lambda_\text{cls}$ and $\lambda_\text{reid}$ are the hyper-parameters to balance the multiple loss terms. Their default values are set to 5, 3, 0.5, respectively.
During training, mask annotations of all VS tasks are fully utilized to train the learnable and prompt queries. 

\textbf{Mask Loss} contains two common functions: Dice loss \cite{dice1945dice} and Binary Cross-Entropy (BCE) loss. It can be formulated as follows:  
$$L_\text{mask} = \sum\nolimits_{t=1}^T L_\text{mask}(M^{t}, \bar{M}^t) + L_\text{mask}(M^{*t}, \bar{M}^t),$$ 
where $M^t, M^{*t}$ are the matched masks for learnable queries and prompt queries, respectively, and $\bar{M}^t$ denotes the ground-truth mask. $t$ and $T$ are the frame index and the number frames of the input video clip.

\textbf{Classification Loss} only applies to category-specified VS tasks. We leverage the similarity between query embeddings and CLIP embeddings of category names for recognition.
The classifier $S$ can be obtained by: 
$$S = 1/T\ \times\ \text{Cosine} (f_\text{cls} ([\mathbf{q}, \mathbf{q^*}]),\ P_{cate}),$$ 
where $P_{cate}$ is the text embedding of category names produced by CLIP text encoder, $f_{cls}$ converts query embeddings from the visual space to the language space using an MLP layer. $T$ is a temperature to amplify the logit. We employ focal loss \cite{lin2017focalloss} to supervise the classifier.

\textbf{ReID Loss}
aims to maintain the temporal consistency in VS tasks, which can be formulated as:
$$L_\text{ReID} =  L_\text{ReID}(\mathbf{q}, \mathbf{q}) + L_\text{ReID}(\mathbf{q}, \mathbf{q}^*) + L_\text{ReID}(\mathbf{q}^*, \mathbf{q}^*),$$ 
where the second term aims to align prompt queries and learnable queries within the same feature space. We utilize the contrastive loss and the auxiliary loss proposed in \cite{pang2021quasi} for the ReID loss. 

\begin{figure*}[h]
     \centering
     \includegraphics[width=0.99\linewidth]{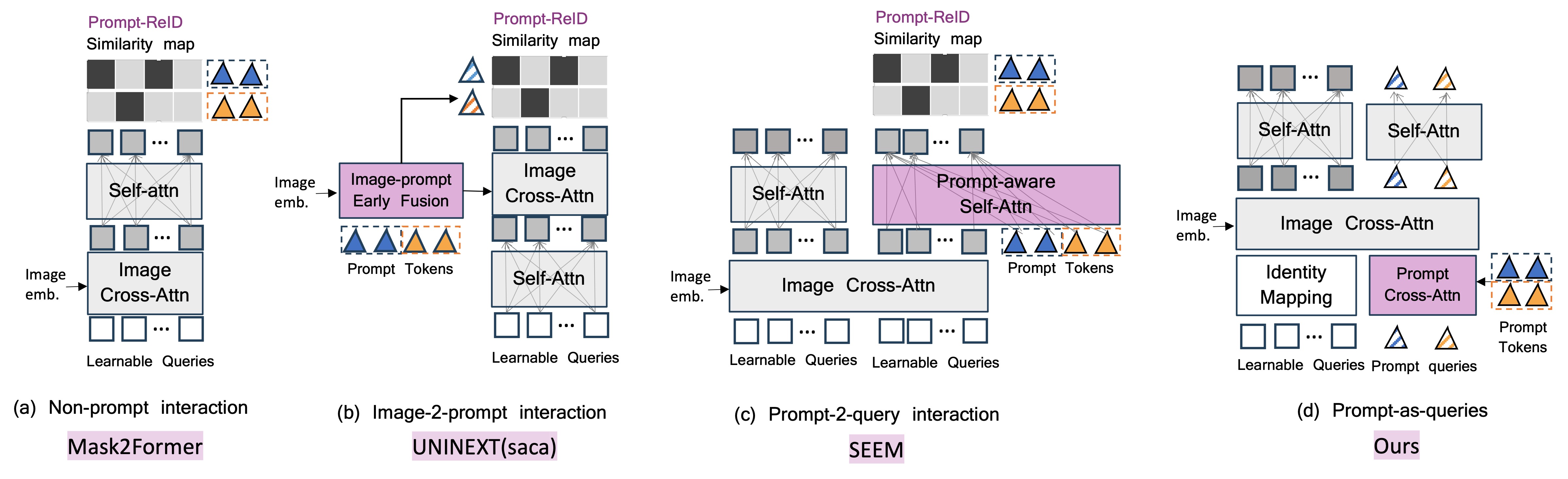}
     \vspace{-2mm}
     \caption{Architecture comparison of mask decoder layers in unified segmentation models, including Mask2Former \cite{cheng2021mask2former}, UNINEXT\cite{uninext}, SEEM\cite{seem} and our UniVS. These methods differ in image and prompt interaction. Note that feed-forward network (FFN) is omitted here.}\label{fig:maskdec}
\end{figure*}
\subsection{Training Stages} 
The whole training process consists of three consecutive stages: image-level joint training, video-level joint training and long video finetuning.
In the first stage, we jointly pretrain UniVS on multiple image datasets, including SA1B \cite{sam}, COCO\cite{lin2014microsoft}, LVIS \cite{gupta2019lvis}, and the mixed dataset of RefCOCO\cite{refcoco}, RefCOCO+\cite{refcoco}, RefCOCOg. Due to limited computational resources, we randomly select 250k images from the 1M images (2.5\%) in the original SA1B dataset for training. It has been experimentally shown in \cite{sam,semanticsam} that the performance of the model trained on 3\% SA1B images is slightly lower than the model trained on the entire images.
In the last two stages, UniVS is trained on image and video datasets, including YT21, OVIS, VIPSeg, YT18, MOSE, Burst and RefYT. Similar to UNINEXT \cite{uninext}, to avoid the model forgetting previously learned knowledge on image-level datasets, we generate pseudo video clips from image datasets and merge them into jointly training on video datasets.

In Table \ref{tab:training_details}, we show the sampling weights of each dataset in each training stage, as well as the number of GPU (GPUs), learning rate (Lr), the maximum iterations (Max Iter) and the time to reduce the learning rate (Step). For UniVS with R50 backbone, the training time for stage 1/2/3 on 16/8/8 V100 GPUs is 9.7/7.5/3.6 days, respectively. And UniVS with Swin-T/B/L backbones need similar training times for stage 1/2/3 on 16/8/8 A100 GPUs. 

\section{More Ablation Studies}
Except specifically stated, experimental results in this section are evaluated using the ResNet50 backbone.

\subsection{Comparison of Unified Architectures}
To show the superiority of our proposed unified video segmentation architecture, we compare UniVS with popular unified segmentation frameworks, including Mask2Former \cite{cheng2021mask2former}, UNINEXT \cite{uninext}, and SEEM \cite{seem}. The architecture comparison is illustrated in Fig. \ref{fig:maskdec}.

The original Mask2Former \cite{cheng2021mask2former} can process multiple category-specified VS tasks, such as VIS/VSS/VPS, but cannot handle prompt-specified segmentation tasks, such as VOS/PVOS/RefVOS. 
UNINEXT \cite{uninext} is an object-centric segmentation model, which aligns text prompts with image embeddings by introducing a vision-language early fusion module in the pixel decoder (see Fig. \ref{fig:maskdec}b). UNINEXT is built upon the DeformabelDETR \cite{zhu2020deformable_detr} framework, which is more suitable for instance-level detection and segmentation, but exhibits relatively weaker performance in detecting and segmenting stuff entities. 
SEEM \cite{seem} is designed for image segmentation. It introduces an extra group of learnable queries and extends the keys and values of self-attention layers to integrate prompt information, as shown in Fig \ref{fig:maskdec}c. However, when multiple prompt entities are presented, SEEM needs to utilize a post-processing matching stage to locate the targets from all predicted masks. 

It can be observed that previous unified architectures require back-end matching between prompt tokens and learnable queries to identify the targets, which is detrimental to maintain entity consistency across frames. In contrast, our UniVS transfers all VS tasks to the prompt-guided target segmentation to explicitly decode masks, and thus the matching strategy is only used when detecting newly appeared entities from learnable queries, as shown in Fig \ref{fig:maskdec}d.

For quantitative performance comparison, we exclude UNINEXT\cite{uninext} here, because it uses DefDETR \cite{zhu2020deformable_detr} architecture instead of Mask2Former architecture, making it hard to be compared directly. We train the Mask2Former, SEEM and UniVS models using the same training settings and datasets (the first two stages in Sec B.2.). The results are shown in Table \ref{tab:abl_maskdec}. While Mask2Former and SEEM  may perform well on some of the VS tasks, UnivS performs the best on almost all VS tasks, demonstrating the superiority of our proposed architecture.
\begin{table}[!t]
\centering
{\small 
\begin{tabular}{p{0.21\linewidth}<{\centering}|p{0.065\linewidth}<{\centering}p{0.065\linewidth}<{\centering}|p{0.055\linewidth}<{\centering}p{0.055\linewidth}<{\centering}|p{0.075\linewidth}<{\centering}|p{0.095\linewidth}<{\centering}}
    \Xhline{0.8pt}
    \rowcolor{isabelline}
    \!\!Video Tasks & \multicolumn{2}{c|}{VIS} &  \multicolumn{2}{c|}{VPS} & \multicolumn{1}{c|}{VOS} & \!\!RefVOS  \\
    \hline
     \multirow{2}{*}{\!\!Method} & {\!\!YT21}  & {\!OVIS}  & \multicolumn{2}{c|}{ VIPSeg} &{ \!\!YT18\ }  & \!\!RefYT \ \\
                             & \!mAP & \!mAP  & \!VPQ & \!STQ & $G^{th}$ & J\&F\\
    \hline
    {\footnotesize \!\!Mask2Former\cite{cheng2021mask2former}}  & 45.9 & 17.2 & 40.1 & 37.9 & - & - \\  
    {\footnotesize \!\!SEEM\cite{seem}}  & 49.2 & 14.7 & 39.3 & 34.2 & 62.1 & * \\
    {\footnotesize \!\!UniVS (Ours) }    & 52.7 & 21.7 & 35.4 & 49.2 & 67.4 & 54.9 \\ 
    \Xhline{0.8pt}
    \end{tabular}
}
\vspace{-1mm}
\caption{\small Quantitative performance comparison among different unified segmentation models. `-' means that the model is inapplicable to this task and `*' means that the result is not reported.
For the VIS task, the results are evaluated on the development set (1/10 of the training set, excluded during training). 
}\label{tab:abl_maskdec}
\vspace{-1mm}
\end{table}

\subsection{Inference Process}
\begin{table}[!t]
\centering
\begin{subfigure}{0.49\textwidth}
    \centering
    \includegraphics[width=0.95\textwidth]{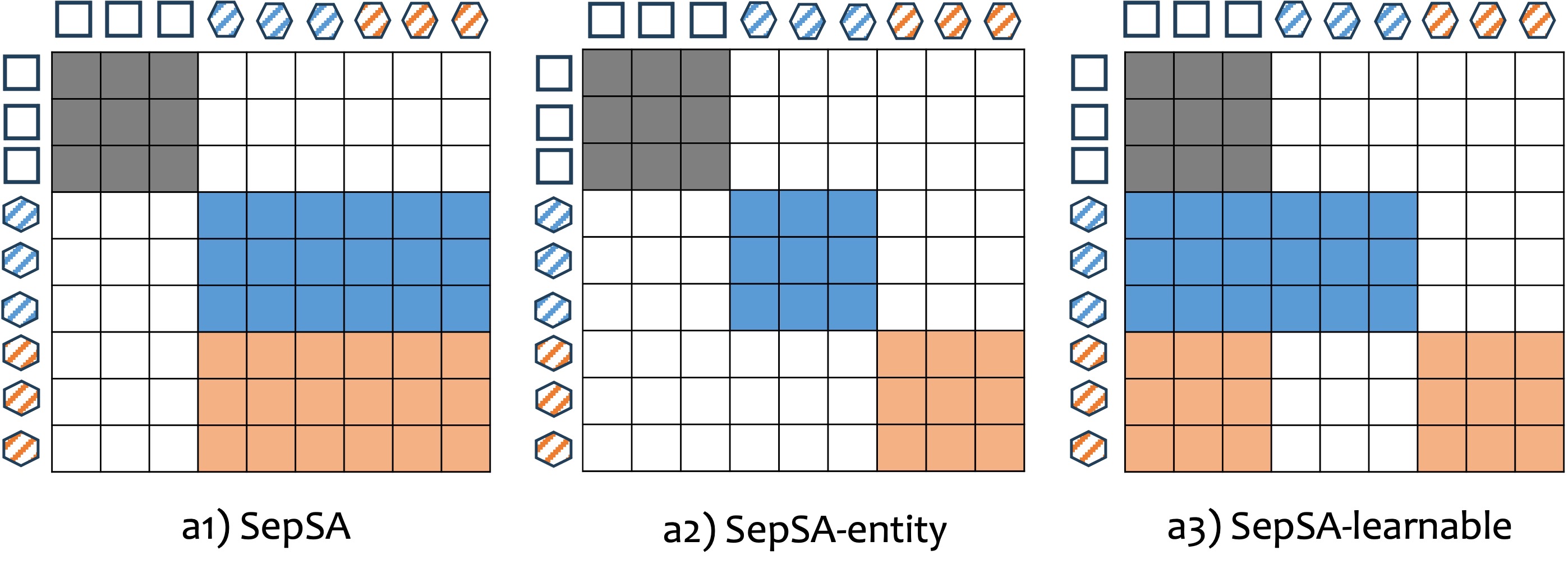}
    \vspace{-1mm}
    \caption{\small Separated self-attention types}
    \vspace{+2mm}
\end{subfigure}
\begin{subtable}{0.49\textwidth}
\centering
\small
\setlength{\tabcolsep}{1.5mm}{
    \linespread{2}
    \begin{tabular}{cccccc}
    \Xhline{0.8pt}
    Self-attn & \multicolumn{5}{c}{PVOS}  \\ 
    \cline{2-6}
    Type & G$^{th \& sf}$ & G$^{th}_{seen}$  & G$^{th}_{unseen}$  & G$^{sf}_{seen}$ & G$^{sf}_{unseen}$ \\
    \Xhline{0.8pt}
    a1 & 61.8 & 59.7 & 57.1 & 68.2 & 62.1 \\
    a2 & 57.9 & 58.4 & 55.0 & 63.5 & 54.5 \\
    a3 & 48.5 & 41.0 & 46.5 & 51.0 & 55.5 \\
    \Xhline{0.8pt}
    \end{tabular}
    }
    \caption{\small Quantitative performance comparison on PVOS task}
    \vspace{+2mm}
\end{subtable}
\begin{subtable}{0.49\textwidth}
\centering
\small
\setlength{\tabcolsep}{2.3mm}{
    \linespread{2}
    \begin{tabular}{ccccc}
    \Xhline{0.8pt}
    Self-attn & \multirow{2}{*}{Used queries} & \multicolumn{3}{c}{RefYTVOS}  \\ 
    \cline{3-5}
    Type &  & J\&F  & J  & F \\
    \Xhline{0.8pt}
    a1 & Prompt  & 38.7 & 36.1 & 41.3 \\
    a2 & Prompt  & 55.7 & 53.9 & 57.5 \\
    a2 & Prompt + Learnable  & 55.1 & 53.5 & 56.8 \\ 
    \Xhline{0.8pt}
    \end{tabular}
    }
    \caption{\small Quantitative performance comparison on RefVOS task}
\end{subtable}
\vspace{-6mm}
\caption{Ablation study on RefVOS tasks, where `SepSA-e' and `SepSA' mean that the separate self-attention mask is executed for each expression and all expressions, respectively. 
} \label{tab:sepsa_refer}
\vspace{-1mm}
\end{table}

\textbf{Separated Self-attention Types.}
As shown in Table \ref{tab:sepsa_refer}(a), in the separated self-attention layer, we test three ways of interaction between learnable and prompt queries. Specifically, 'SepSA' refers to separate self-attention calculations for learnable and prompt queries respectively. 'SepSA-entity' involves interaction among prompt queries belonging to the same entity, with no visibility across different entities. Lastly, 'SepSA-learnable' builds upon 'SepSA-entity' by allowing each prompt query to see all learnable queries to extract the overall image information.

To evaluate the impact of these three approaches on visual prompt-guided video segmentation tasks, we conducted an ablation study on the PVOS task, which involves simultaneous thing and stuff object segmentation. As shown in Table \ref{tab:sepsa_refer}(b), the experimental results demonstrate that 'SepSA' performs the best, as it avoids content overflow between prompt and learnable queries.

\textbf{Efficient Inference on Prompt-guided Segmentation.}
As shown in Table \ref{tab:sepsa_refer}(c), UniVS can simultaneously process multiple prompt-guided targets in the RefVOS task by applying entity-wise separated self-attn mask (termed as SepSA-entity). This inference process is more efficient than the existing methods that often segment targets one by one. Additionally, using only prompt queries can achieve higher performance than using both prompt and learnable queries.

\textbf{The Detection of Newly Appeared Objects.}
For category-specified VS tasks, we investigate the impact of using different interval frames on detecting newly appeared objects. Since the VSS task only requires pixel-level category prediction without the need of instance-level tracking, it does not detect new objects. Therefore, we conduct ablation on the VIS and VPS tasks. The results are shown in Table \ref{tab:abl_newly}.
It can be observed that when the number of interval frames is smaller than the number of frames (\ie, 5) in the input video clip, the performance is basically unaffected. However, if the interval frames exceed the number of frames in the input video clip, the performance is decreased by 1$\sim$2\%. This decline can be attributed to the missing of some newly appeared objects.
\begin{table}[!t]
\centering
\setlength{\tabcolsep}{2.5mm}{
\begin{tabular}{c|cc|cc}
    \Xhline{0.8pt}
    \rowcolor{isabelline}
     Video Tasks & \multicolumn{2}{c|}{VIS} &  \multicolumn{2}{c}{VPS}  \\
    \hline
     \multirow{2}{*}{Interval frames} & {\!\!YT21}  & {\!OVIS}  & \multicolumn{2}{c}{ VIPSeg} \\
        & \!mAP & \!mAP  & \!VPQ & \!STQ \\
    \hline
    1 & 54.8 & 24.2 & 38.3 & 46.2 \\
    3 & 54.6 & 23.7 & 38.6 & 45.8 \\
    5 & 54.6 & 23.4 & 38.4 & 45.8 \\
    7 & 53.0 & 22.1 & 38.2 & 45.2 \\
    9 & 52.6 & 22.0 & 37.7 & 45.1 \\
    \Xhline{0.8pt}
    \end{tabular}
}
\vspace{-1mm}
\caption{\small Ablation study on the number of interval frames to detect newly appeared objects. The input clips include 5 frames. 
}\label{tab:abl_newly}
\end{table}

\textbf{Inference speed.} Table \ref{tab:inf_speed} shows the inference speed of UniVS with 640p video as input. Videos in YT21, YT18 and RefYT contain 1 $\sim$ 3 objects, whereas videos in VSPW, VIPSeg and VIPOSeg have more than 15 entities, whose inference speed is slower.
\begin{table}[!tbb]
\vspace{-2mm}
\centering
{\small
\setlength{\tabcolsep}{1.2mm}{
      \linespread{2}
      \begin{tabular}{l|ccc|ccc}
         \Xhline{1pt}
         Task    & VIS  & VSS  & VPS    & VOS  & RefVOS & PVOS \\
         Dataset & YT21 & VSPW & VIPSeg & YT18 & RefYT  & VIPOSeg \\
         \hline
         FPS     & 20.2 & 15.3 & 10.4   & 17.5 & 20.0  & 11.9 \\
         \Xhline{1pt}
      \end{tabular}
}}
\vspace{-2mm}
\caption{Inference speed of UniVS with ResNet50 backbone on a single V100 GPU.
}\label{tab:inf_speed}
\end{table}

\subsection{Generalization Ability} 
To further verify the generalization capability of UniVS, we try to train UniVS solely on the category-guided VS datasets but test on the prompt-guided VS datasets. 
In Table \ref{tab:cate2prompt}, we train UniVS only on category-specific VS tasks, including COCO, LVIS, YT21, OVIS and VIPSeg datasets. The testing is conducted on two prompt-guided VS tasks: VOS and PVOS. Experimental results demonstrate that UniVS exhibits comparable or even better performance on VOS and PVOS tasks, indicating its remarkable generalization ability. Additionally, we speculate that the significant performance improvement on DAVIS is due to its similarity to the training data distribution, while the slight performance drop on VIPOSeg is attributed to its inclusion of more diverse video scenes and objects, which exceeds the distribution of training data.

\begin{table}[!thb]
\centering
{\small
\setlength{\tabcolsep}{1.mm}{
      \linespread{2}
      \begin{tabular}{cc|c|ccccc}
         \Xhline{1pt}
         \multirow{2}{*}{Training } & \multirow{2}{*}{data} & VOS & \multicolumn{5}{c}{PVOS} \\
         & & DAVIS & \multicolumn{5}{c}{VIPOSeg} \\
         Category & Prompt  & G$^{\text{th}}$ & G$^{\text{th}}_{\text{seen}}$ & G$^{\text{th}}_{\text{unsn}}$ & G$^{\text{sf}}_{\text{seen}}$ & G$^{\text{sf}}_{\text{unsn}}$ & G$^{\text{th\& sf}}$ \\
         \Xhline{1pt}
         $\checkmark$ & $\checkmark$ & 70.8 & 63.4 & 61.9 & 73.9 & 68.4 & 66.8 \\
         $\checkmark$ & \crossmark   & 75.0 & 59.2 & 54.2 & 67.9 & 78.1 & 64.9 \\
         \Xhline{1pt}
      \end{tabular}
}}
\vspace{-2mm}
\caption{\small Generalization ability of UniVS trained on category-guided VS tasks but tested on prompt-guided VS tasks. UniVS adopts SwinB backbone and trained on stages 1\&2.
}\label{tab:cate2prompt}
\vspace{-5mm}
\end{table}


\section{Visualization}
\begin{figure*}[t]
     \vspace{-1mm}
     \centering
     \includegraphics[width=1\linewidth]{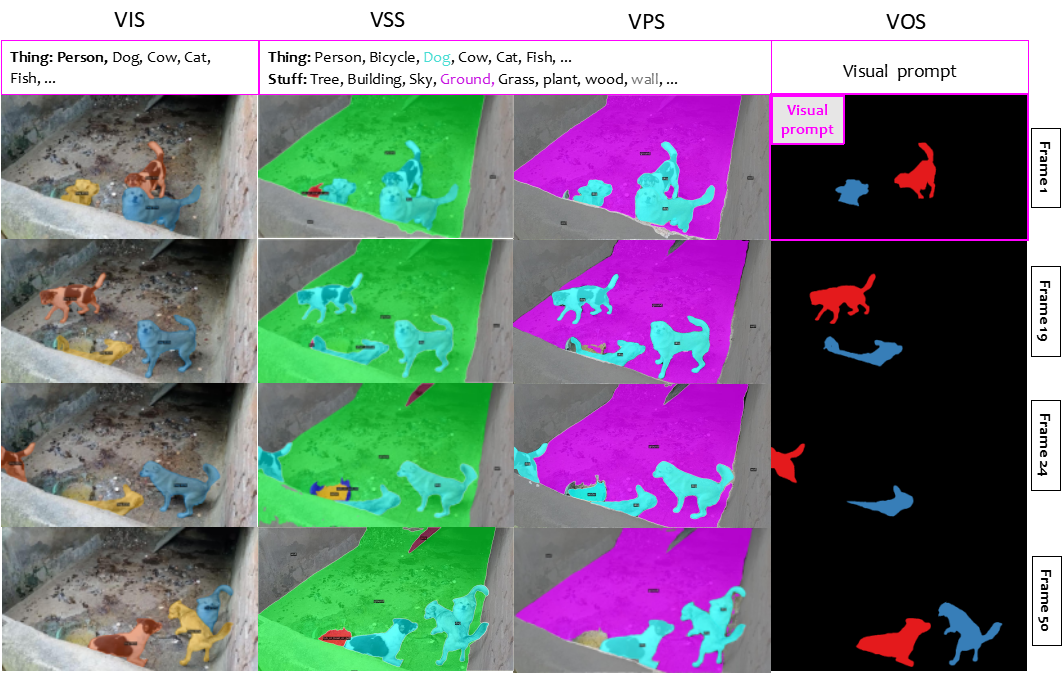}
     \includegraphics[width=1\linewidth]{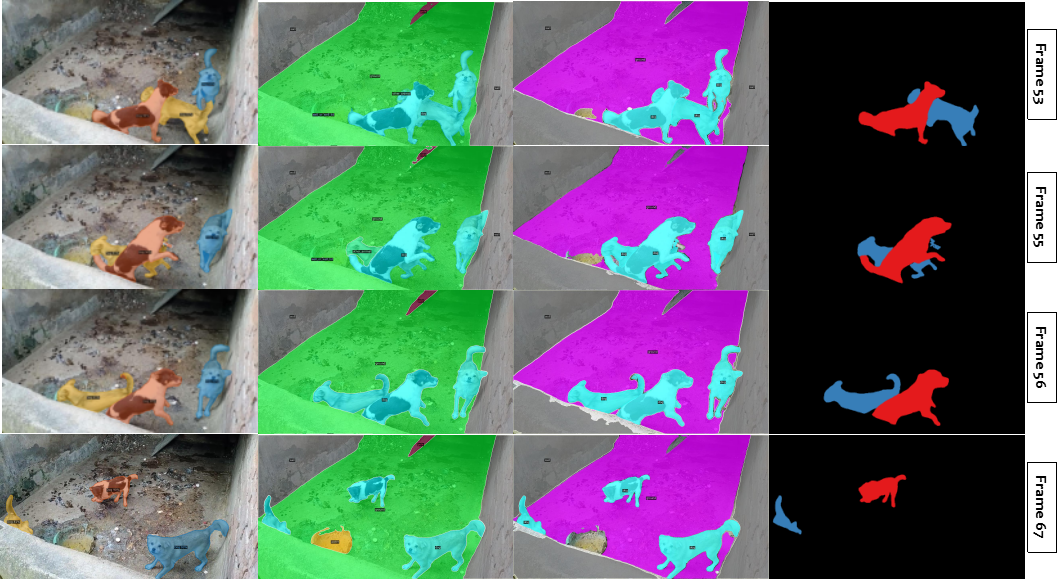}
     \vspace{-5mm}
     \caption{Visualization examples of UniVS on \textbf{VIS/VSS/VPS/VOS} tasks. The original videos come from the validation set of OVIS dataset, while the entity categories of VIS and VSS/VPS are from OVIS and VIPSeg datasets, respectively. The visual prompts are from the MOSE dataset.
     }\label{fig:supp_vs_tasks1}
     \vspace{-2mm}
\end{figure*}

\begin{figure*}[t]
     \vspace{-1mm}
     \centering
     \includegraphics[width=1\linewidth]{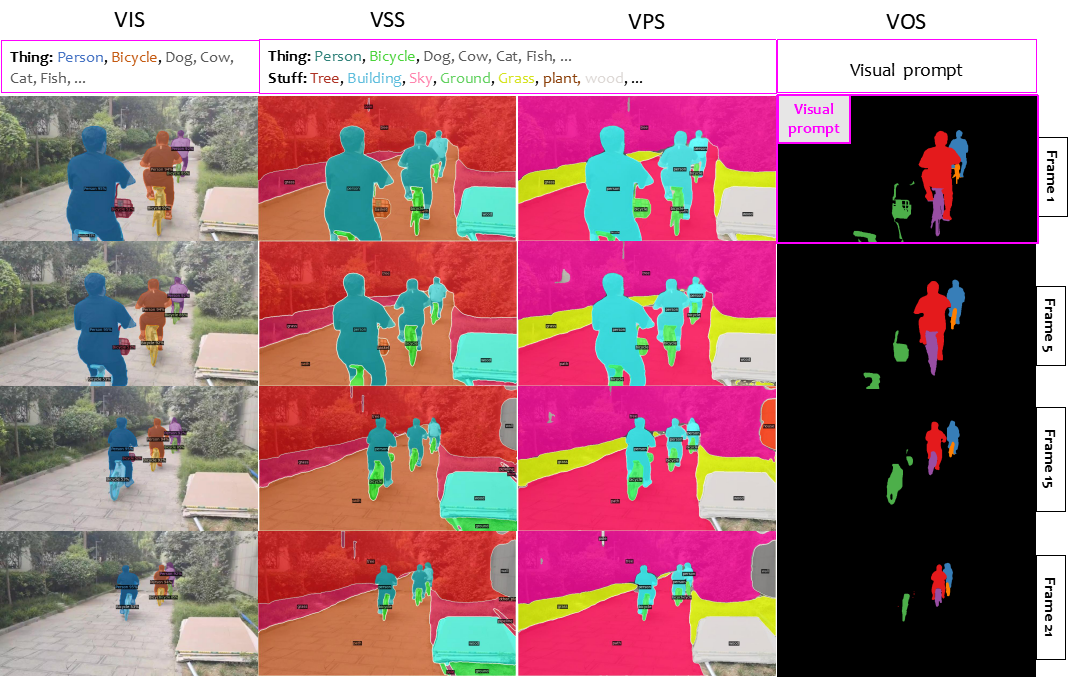}
     \includegraphics[width=1\linewidth]{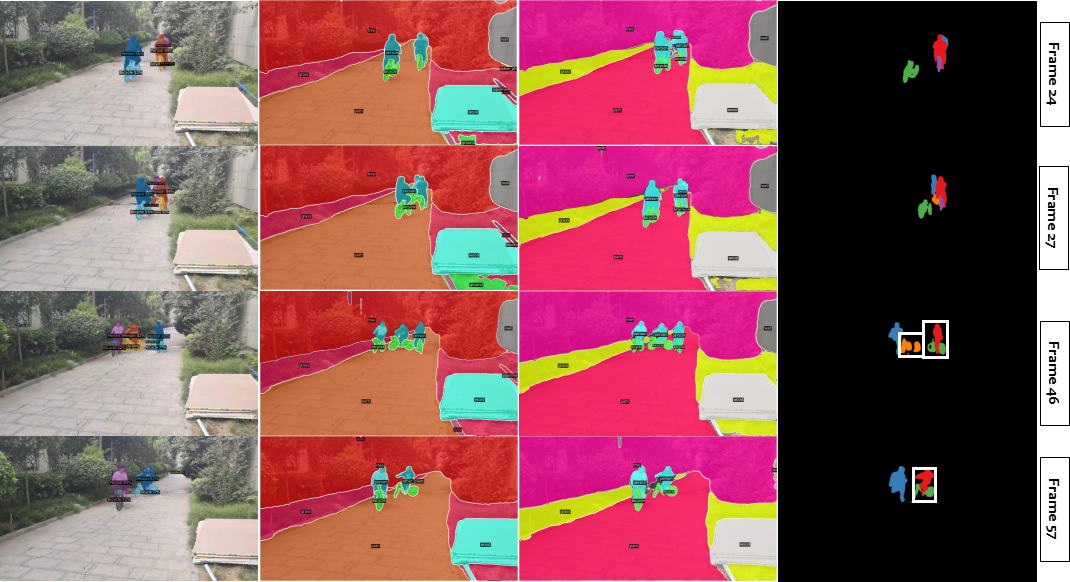}
     \vspace{-5mm}
     \caption{Visualization examples of UniVS on \textbf{VIS/VSS/VPS/VOS} tasks. The original videos come from the validation set of OVIS dataset, while the entity categories of VIS and VSS/VPS are from OVIS and VIPSeg datasets, respectively. The visual prompts are from the MOSE dataset. For VOS task, we mark the incorrectly tracked objects in the last column with white bounding boxes.
     }\label{fig:supp_vs_tasks2}
     \vspace{-2mm}
\end{figure*}

\textbf{VIS/VSS/VPS/VOS.}
Figs. \ref{fig:supp_vs_tasks1} and \ref{fig:supp_vs_tasks2} display the segmentation results predicted by our UniVS on VIS/VSS/VPS/VOS tasks. To enhance visualization, we use the same video for different VS tasks. Specifically, the thing categories for the VIS task is sourced from the OVIS dataset, while the thing and stuff categories for the VSS and VPS tasks are derived from the VIPSeg dataset. As for the VOS task, the visual prompts are obtained from the MOSE dataset. It can be observed that UniVS achieves satisfactory segmentation results across these tasks, demonstrating its excellent generalization capability.

\begin{figure*}[t]
     \vspace{-1mm}
     \includegraphics[width=0.95\linewidth]{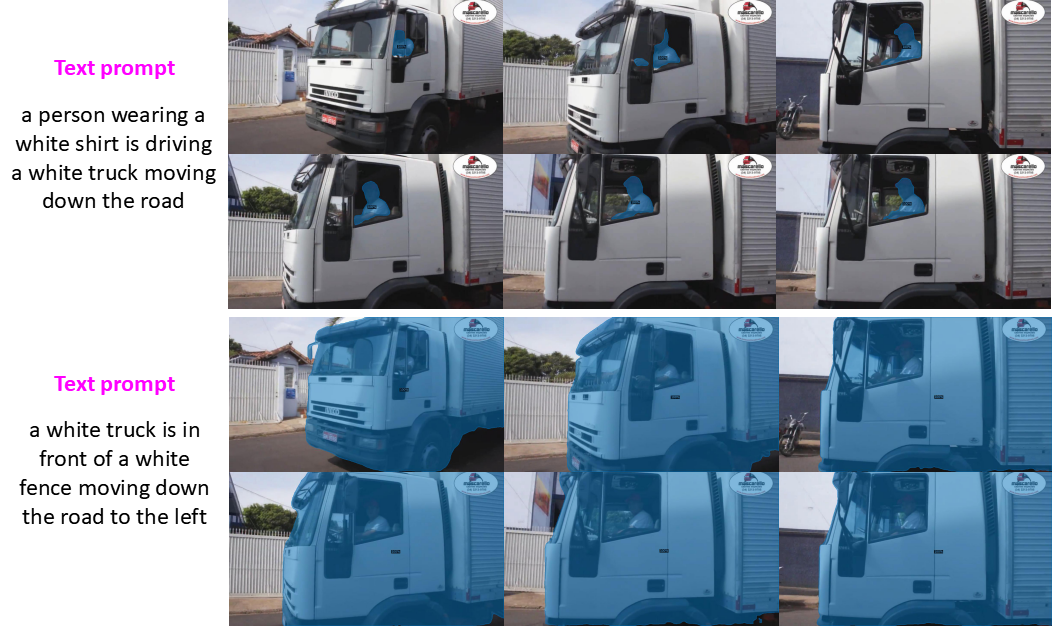} 
     \includegraphics[width=0.95\linewidth]{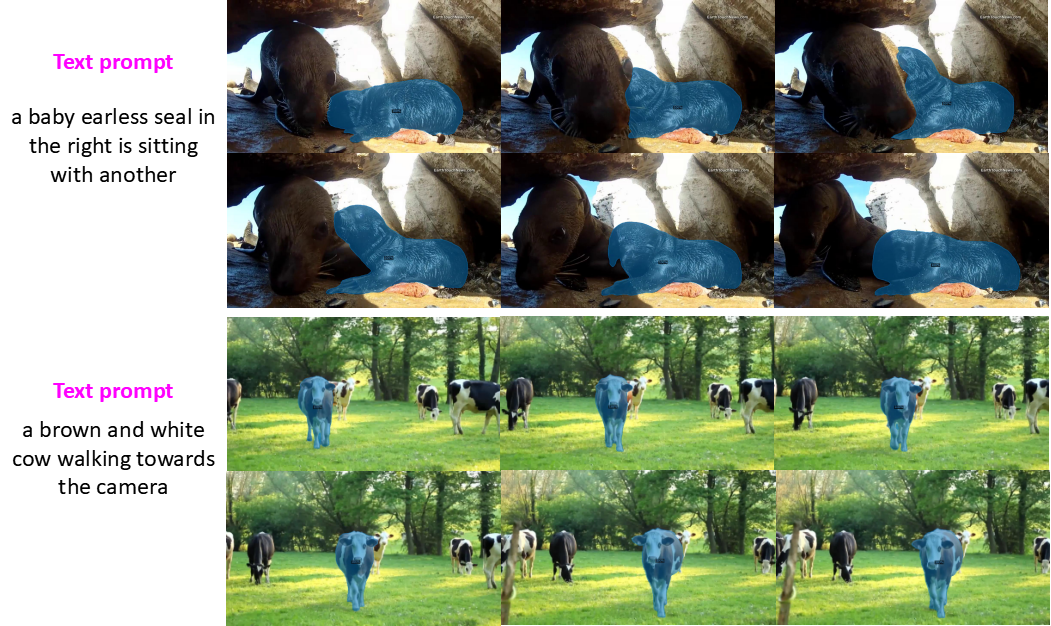}
     \vspace{-1mm}
     \caption{Visualization examples of UniVS with \textbf{text prompts} in the \textbf{RefVOS} task. The videos are from the RefYTVOS valid set, and the left side provides the expression per object.
     }\label{fig:supp_refer}
     \vspace{-1mm}
\end{figure*}

\textbf{RefVOS.} 
Fig. \ref{fig:supp_refer} exhibits the video segmentation results with text expressions as prompts. We observe that UniVS can accurately segment objects in the video based on the given text prompts. This demonstrates that UniVS can effectively integrate language and video information, enabling cross-modal consistent segmentation.

\begin{figure*}[t]
     \vspace{-1mm}
     \centering
     \includegraphics[width=0.825\linewidth]{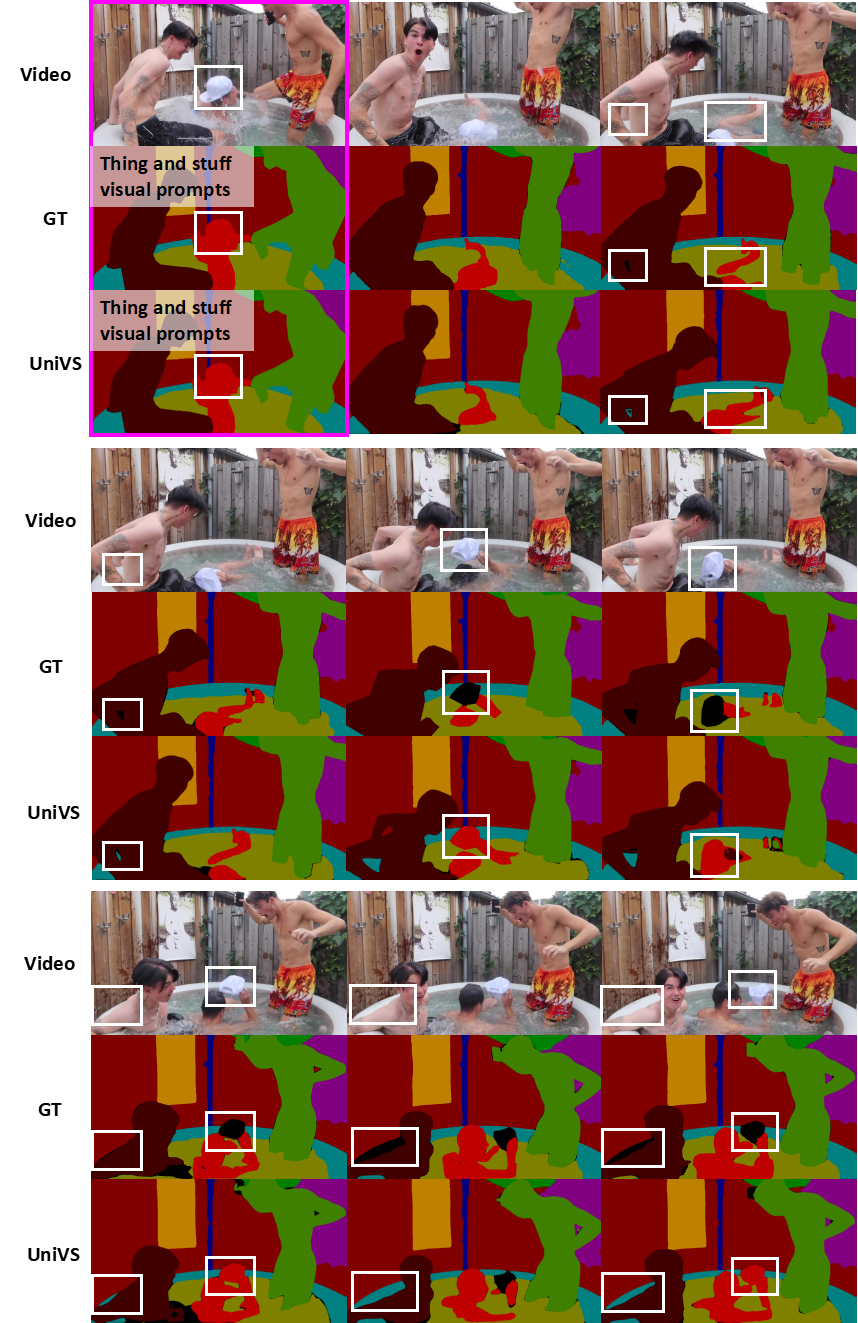} 
     \vspace{-1mm}
     \caption{Visualization examples of UniVS with \textbf{visual prompts} in the \textbf{PVOS} task. The video frames are from the VIPOSeg valid set, with the second row showing the ground truth masks and the last row displaying the predicted masks by our UniVS. Note that the visual prompts include both thing and stuff classes.
     }\label{fig:supp_pvos}
     \vspace{-1mm}
\end{figure*}

\textbf{PVOS.} Fig. \ref{fig:supp_pvos} displays the segmentation results for the PVOS task. The second and third rows compare the ground truth masks with the UniVS predicted masks, confirming the superiority of UniVS in visual prompt-guided thing and stuff entity segmentation.
Additionally, it is worth noting that due to the high cost of video segmentation annotation, this dataset adopts a semi-automatic annotation approach, combining manual and algorithmic annotations. This may result in potential omissions or inaccuracies in the provided ground truth masks, such as the areas highlighted by the white bounding boxes. Therefore, UniVS also holds potential as a complementary method for dataset annotation in future endeavors.

In summary, UniVS demonstrates excellent general segmentation capability and can handle various VS tasks. It is not only suitable for category-guided segmentation but also performs well in almost all visual prompt-guided thing and stuff entity segmentation tasks. Meanwhile, UniVS showcases its ability in expression-guided cross-modal object segmentation tasks. Its multi-modal fusion capability and consistent segmentation performance make UniVS highly promising for integrating language and video information.

\textbf{Video Demo.} We provide more visualizations of the segmentation results on the project page. Please access the related content by clicking on the link \url{https://sites.google.com/view/unified-video-seg-univs}.


\clearpage 


\end{document}